%% file: IROS2016.tex
\documentclass[letterpaper, 10 pt, conference]{ieeeconf}  

\IEEEoverridecommandlockouts                              

\usepackage{graphics, graphicx} 
\usepackage{epsfig} 
\usepackage{mathptmx} 
\usepackage{times} 
\usepackage{amsmath} 
\usepackage{amssymb}  
\usepackage{color}
\usepackage{algorithmic}
\usepackage{algorithm}
\usepackage{stfloats}
\usepackage{subfigure}
\usepackage{setspace}
\usepackage{import}
\usepackage{breqn}
\usepackage{gensymb}

\newcommand{\vtxt}[1]{}

\title{\LARGE \bf
Visual Servoing in Orchard Settings}

\author{Nicolai H\"ani$^{1}$ and Volkan Isler$^{2}$
\thanks{$^{1}$Nicolai H\"ani is with Zurich University of Applied Sciences and currently a visiting scholar at University of Minnesota,
       Minneapolis, MN 55455, USA,
        {\tt\small haeni001@umn.edu}}%
\thanks{$^{2}$Volkan Isler is with the Department of Computer Science and Engineering,  University of Minnesota,
        Minneapolis, MN 55455, USA
        {\tt\small isler@cs.umn.edu}}%
}

\begin{document}
\singlespace
\maketitle
\thispagestyle{empty}
\pagestyle{empty}

\begin{abstract}
\input{abs}
\end{abstract}

\input{intro}
\input{relwork}
\input{approach}
\input{initial}
\input{vision}
\input{expt}
\input{conc}

\section{Acknowledgements}
We thank UMN Department of Horticultural Science for allowing us to perform experiments in their orchard.
This work was supported in part by UMN MnDrive initiative,  NSF grants \# 1317788  and \# 1111638, and USDA NIFA MIN-98-G02.

\bibliographystyle{IEEEtran}
\bibliography{refs}

\end{document}

%% file: abs.tex
We present a general framework for accurate positioning of sensors and end effectors in farm settings using a camera mounted on a robotic manipulator. Our main contribution is a visual servoing approach based on a new and robust feature tracking algorithm. Results from field experiments performed at an apple orchard demonstrate that our approach converges to a given termination criterion even under environmental influences such as strong winds, varying illumination conditions and partial occlusion of the target object. 
Further, we show experimentally that the system converges to the desired view for a wide range of initial conditions. This approach opens possibilities for new applications such as automated fruit inspection, fruit picking or precise pesticide application.



%% file: intro.tex
\section{INTRODUCTION}

There is significant interest in automating farm tasks driven by increasing demand for food, insufficient labor availability and widespread use of precision agriculture methods. As a result, various robots and other machinery are being developed for weeding, picking, fertilizer or pesticide application and similar tasks. Performing these in an autonomous fashion in unstructured environments such as orchards is a challenging next frontier for robotics.

In this paper we focus on the development of an autonomous system for close up apple inspection. Suppose a robot has identified a fruit in its field of view and must place an end effector or a sensor to a given pose with respect to the fruit. For example, many apple diseases start at the stem or the blossom ends of the fruit, which are not easily observable from an outside view \cite{_apple_????}. Instead, a camera mounted on a manipulator can be used to perform close-up inspection of the fruits. Another interesting field of application is automated acquisition of empirical data. Various works have used Near Infra-Red (NIR) technology to measure parameters such as the content of fructose, glucose, the internal moisture of the fruit and their acidity \cite{alander_review_2013}. While these tasks are currently performed either manually or automated after the picking of a fruit, a fully autonomic framework which allows for inspection in the field throughout the growing process is desirable. The resulting sensor placement task resembles a classical visual servoing problem for which there exist two primary approaches \cite{malis_survey_2002}:

\begin{enumerate}
\item Image Based Visual Servoing (IBVS) approaches use observations of distinctive features and their desired final locations in the image plane to compute the camera movement. Common features used are point features or more general geometric features such as lines and ellipses.
\item Position Based Visual Servoing (PBVS) approaches estimate the object's pose using an accurate 3D model of the object. The necessary camera movement is computed using the comparison between the object pose and a desired view pose of the camera.
\end{enumerate}

\begin{figure}[htb]
	\centering
	\includegraphics[width=0.4\columnwidth]{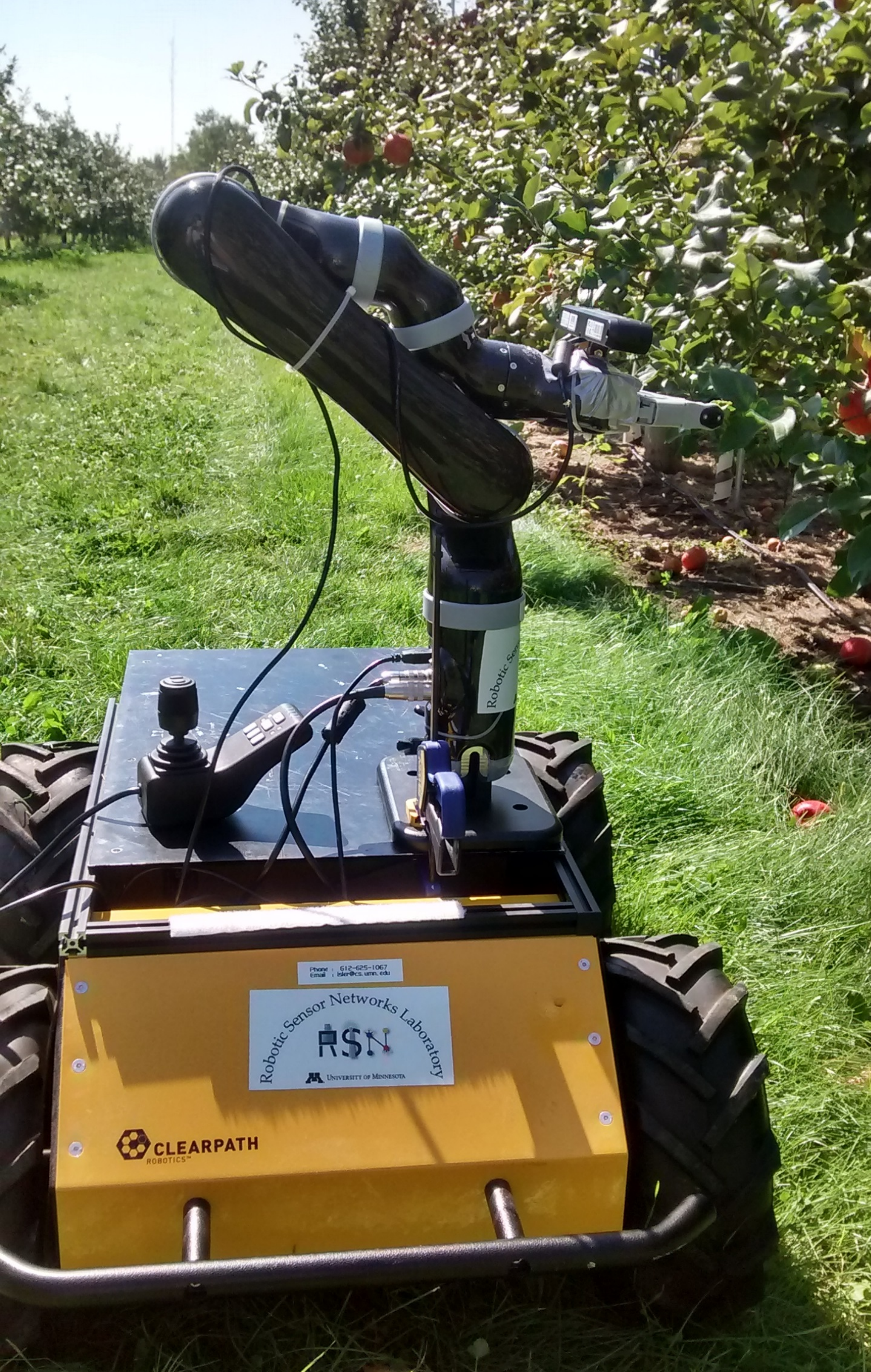}
	\caption{Robotic system for close up inspection in apple orchard setting}
	\label{fig:systemSetup}
\end{figure}

In the case of outdoor apple fruit inspection, both approaches have limitations. The PBVS approach is well known to have stability issues when calibration errors or errors in the computed pose are introduced into the system \cite{chaumette_potential_1998}. The variability of apple shapes as well as the presence of environmental conditions such as inconsistent illumination and/or partial occlusion of the target make this approach unsuitable for our setup. IBVS approaches on the other hand have stability issues when the image Jacobian becomes singular as we will see later in this paper. We will further see how this issue can be overcome by the correct choice of image features. Further it is challenging to define IBVS features a priori for a wide variety of shapes and to track them as their visibility and illumination changes. \emph{Our primary contribution is an approach to overcome these challenges.}

Our work is based on the assumption that in a previous step a partial or complete model of the environment was acquired, and the target fruits have been identified. Further, we assume that the target pose is within the reach of the manipulator. One justification of this assumption is that approximate locations of fruits can be identified in a surveying step e.g., using Unmanned Aerial Vehicles (UAV)~\cite{roy2015apple} which can be used to {\em roughly} place the base near the fruit. The focus of this paper is on the following step of {\em precisely} placing the end effector with respect to the identified fruit.
We start the paper by reviewing related  work, after which we present an overview of our system. Section~\ref{sec:approach} will introduce the underlying geometric model and justify our selection of the feature vector. Afterwards we present in Section~\ref{sec:initial} the computation of the initial position together with our choice of servo controller (which was developed by F: Chaumette in~\cite{chaumette_image_2004}). Our main contributions, namely the robust tracking approach and experimental validation both in controlled indoor and outdoor settings are presented in Sections~\ref{sec:controller} and~\ref{sec:exp}.

%% file: relwork.tex
\section{Related Work}
\label{sec:previous}

Vision based manipulator control, commonly known as visual servoing, is an established research area. Early approaches date back to 1979 when Hill and Park introduced the terminus system~\cite{hill_real_1979}. 
Today, well-established methods are routinely used in controlled settings such as factory floors~\cite{hutchinson1996tutorial}. Extending these approaches to dynamic environments with inconsistent or changing illumination and occlusions remains an active research domain.

Recent work by Panday et al.~\cite{pandya_servoing_2015} uses a linear combination of 3D models  to provide visual servoing for various object categories. This approach uses a pose estimation which is possible due to the unambiguous nature of the models, the availability of accurate 3D model instances and a setup which offers consistent illumination and does not contain clutter or occlusion. Kurashiki et al.~\cite{kurashiki_visual_2015} used an IBVS approach to control a mobile robot to follow a road boundary. Although the approach has been tested in indoor and outdoor environments so are the extracted features only used to enhance the scalability of additionally acquired LIDAR data. Other proposed approaches used visual servoing techniques to completely or partially control UAV's in outdoor environments. Mondragon et al.~\cite{mondragon_3d_2011} implemented a 3 Degree Of Freedom (DOF) controller for continuous following of a 3D moving target by a UAV. The approach is robust to illumination changes and dynamic enough to filter wind disturbances. The single target however is rather large and clearly distinguishable from the background. In our approach we go a step further by providing object tracking in cluttered, occluded environments where multiple, small targets are present in the images. 
Mahony et al.~\cite{mahony_image-based_2005} track parallel lines to maneuver a UAV. They provide a controller designed specifically for the UAV's kinematic setup together with simulations but unfortunately no analysis of the controllers behavior in outdoor environments. 
I. Sa and P. Corke~\cite{sa_improved_2013} used a line tracker for tracking of a pole in indoor and outdoor environments. In outdoor environments an IMU sensor is used for model prediction and they compare their tracker to the moving edges tracker implemented in the VISP library~\cite{Marchand05b}. Although their approach beats the VISP tracker this comes as no surprise as they are able to rely on the IMU data for model prediction that allows for an active recovering after dropped frames. In our approach we only use the image data to control a robot and we provide a mechanism for recovering solely from visual data.
None of the mentioned approaches take occlusions into consideration and the controllers are only used in non-, or semi-cluttered environments. With our proposed solution we overcome these limitations by introducing appropriate features, an approach for tracking them and  experiments to validate our approach in field conditions.

%% file: approach.tex
\section{Overview of the Approach}
\label{sec:approach}
\begin{figure}[ht!]
	\centering
	\includegraphics[width=\columnwidth]{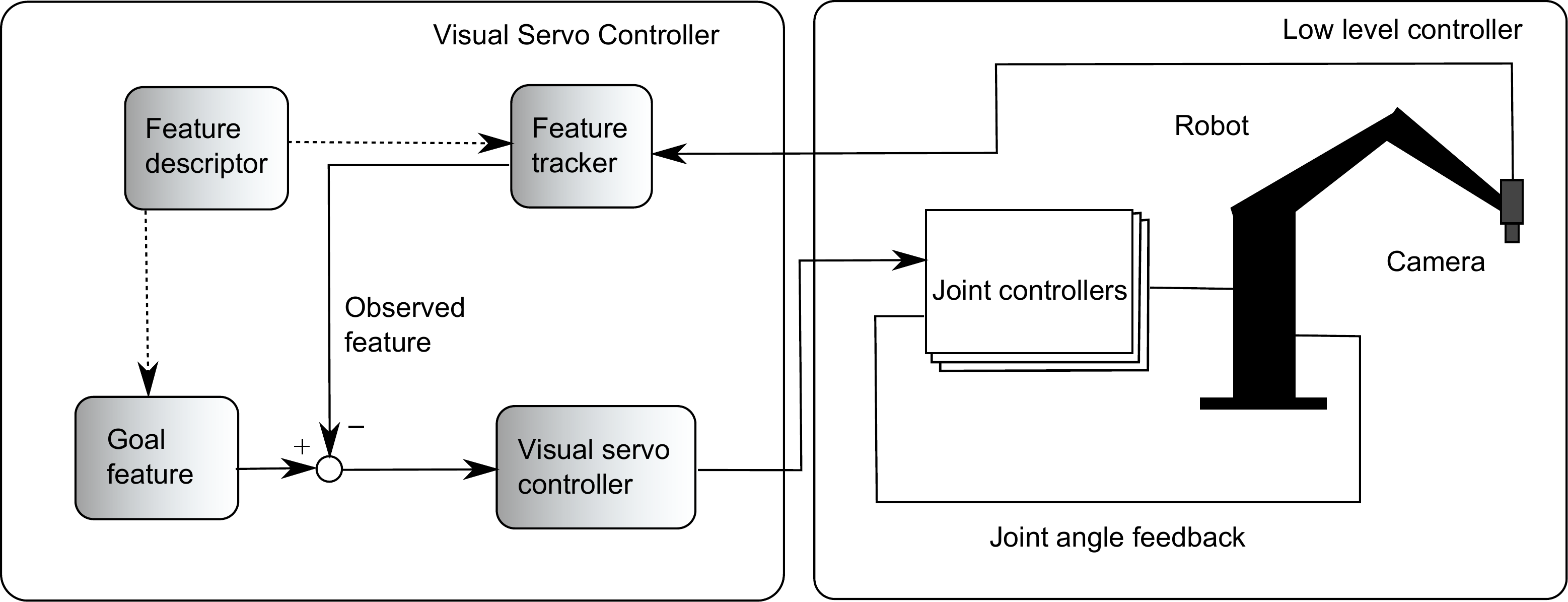}
	\caption{Overview of our approach}
	\label{fig:flowchart}
\end{figure}

A schematic overview of our approach is given in Figure~\ref{fig:flowchart}. We assume the availability of a preprocessing step that places the robot base near the fruit. The following visual servo control loop relies on the detection and tracking of features.

\subsection{System Description} \label{sec:system}
The system setup, shown in Figure~\ref{fig:systemSetup},  consists of a 6 DOF Kinova Mico robotic manipulator and an Asus Xtion sensor which is rigidly mounted on the robots end effector. The system can be mounted on an Unmanned Ground Vehicle (UGV) such as the Husky developed by Clearpath robotics.
While the depth information from the sensor is valuable, {\em the IBVS approach presented here relies only on the RGB component of the input images.} There are two reasons for the choice of using RGB data as our control input. (i)~The main reasons is that acquisition of depth data by sensors such as the Asus Xtion or the Microsoft Kinect requires the projection of a light pattern in the NIR spectrum (around 800 nm) onto the environment. By observing this pattern with a special NIR camera the depth information can be computed by comparing the observed pattern to an undistorted one. In the presence of strong illumination sources, which is typical in outdoor settings, the depth information is not available as the sun superimposes the laser information. (ii)~The Asus Xtion in our case requires a minimum sensing distance of approximately $0.8 \mathrm{m}$ which makes it inappropriate for some close-up inspection tasks.
However, we use the RGB-D platform for this project because it offers an inexpensive color camera with a resolution of $640\times480 \mathrm{pixel}$ at $30 \mathrm{fps}$. Due to its lightweight construction, it is ideal for an eye in hand configuration.

Although IBVS approaches are known to be robust with respect to calibration errors as well as errors in the computation of the feature depth $Z$~\cite{yoshikawa_effect_1994} our approach depends on the transformation of the camera velocities into robot velocities. For this purpose, we used the hand-eye calibration method proposed by Tsai and Lenz~\cite{tsai_new_1989} to calibrate our system.

\subsection{Low level controls}
The low level controls were provided by the manufacturer and executed on an embedded controller inside the manipulator. It consists of a PID controller and a ROS interface that allows us to control the robot using position or velocity commands in Cartesian or joint space. This design has one major drawback: during the testing stage it turned out that the velocity  controller puts several thresholds on the inputs. Any velocity that falls under this threshold is simply filtered out by the controller. In our visual servoing approach the resulting camera velocities become small as we get closer to the target. We overcome this limitation by using an adaptive-, instead of a constant gain. This adaptive gain increases the velocities incrementally when the feature error becomes small.

\subsection{High level controls}
The high level control framework is based on modular software packages we developed: a hardware abstraction of the robot is used as its software representation. It keeps the internal states up to date and communicates with the hardware. A tracker instance is responsible for the visual feature detection. The features are handed over to the visual feedback controller which generates the camera velocities using a visual servoing control law. Using the hardware abstraction and the computed robot Jacobian we compute the resulting robot velocities by transforming the camera velocities into the robot coordinate frame with

\begin{equation}
^rv_r = \begin{bmatrix}
^rR_c & S(^rT_c)^rR_c \\
0_{_{3x3}} & ^rR_c
\end{bmatrix}_{_{6x6}}{^cv_c}
\end{equation}

where $r$ denotes the robots-, and $c$ the camera's coordinate system. $S$ is the skew symmetric matrix of the angular velocities $\omega(t)$, $R$ is the rotation and $T$ the translation between those coordinate systems. The states of the rotation and translation are updated at every time step $t$ and kept in the hardware abstraction layer. The visual servoing control law was implemented using the VIsual Servoing Platform (VISP) from the Lagadic Research Laboratory in Rennes, France~\cite{Marchand05b}. 

\subsection{Modeling of the Feature Vector}
Providing automated visual inspection of apple fruits enforces constrains on the selection of the visual feature vector, namely:

\begin{itemize}
\item The have to provide valid and consistent features for a variety of object shapes and textures.
\item The features have to lead to a stable control law.
\item The chosen features have to be robust with respect to illumination changes and partial occlusion of the target.
\end{itemize}

Point features, although well established in visual servoing research, can not be used in our setup as no consistent apple shapes and textures are present. 
A position based approach is not taken into consideration as it requires accurate, unambiguous 3D models and an accurate calibration method. 

To define a feature vector for a variety of shapes we approximate the apple shape by a sphere and use its projection into the image plane as our visual feature. The projection of a sphere in the image plane is an ellipse or a circle in the case where the sphere and the camera's optical axis align. However experiments using elliptical features that were manually initialized and tracked using a moving edges tracker (MET) showed that illumination changes and occlusion lead to holes and deformations of the boundary. 

\begin{figure}[hbt]
\centering     
\subfigure[Ellipse in image $I_k$]{\label{fig:ellipse1}\includegraphics[width=0.45\columnwidth]{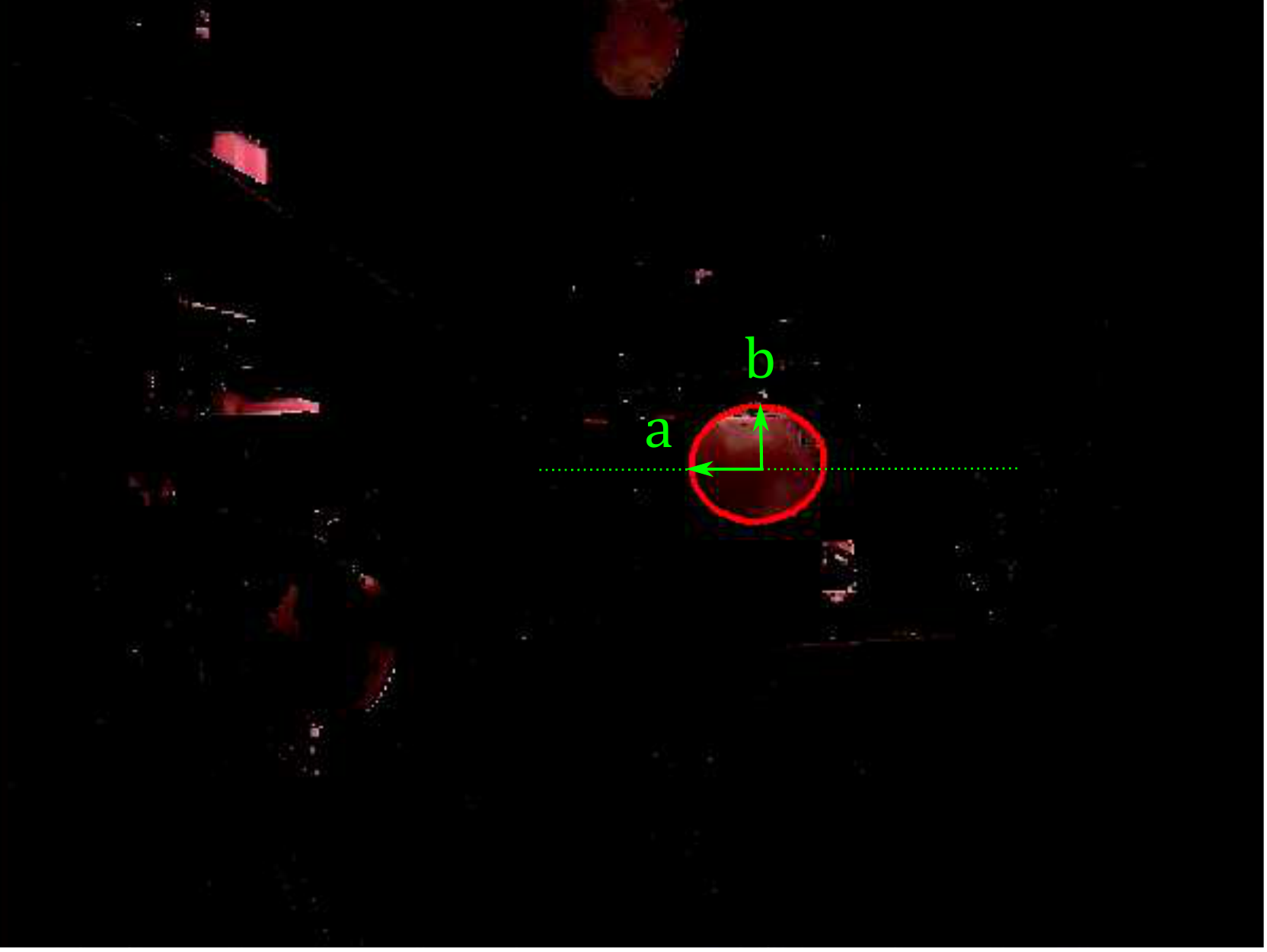}}
\subfigure[Ellipse in image $I_{k+1}$]{\label{fig:ellipse2}\includegraphics[width=0.45\columnwidth]{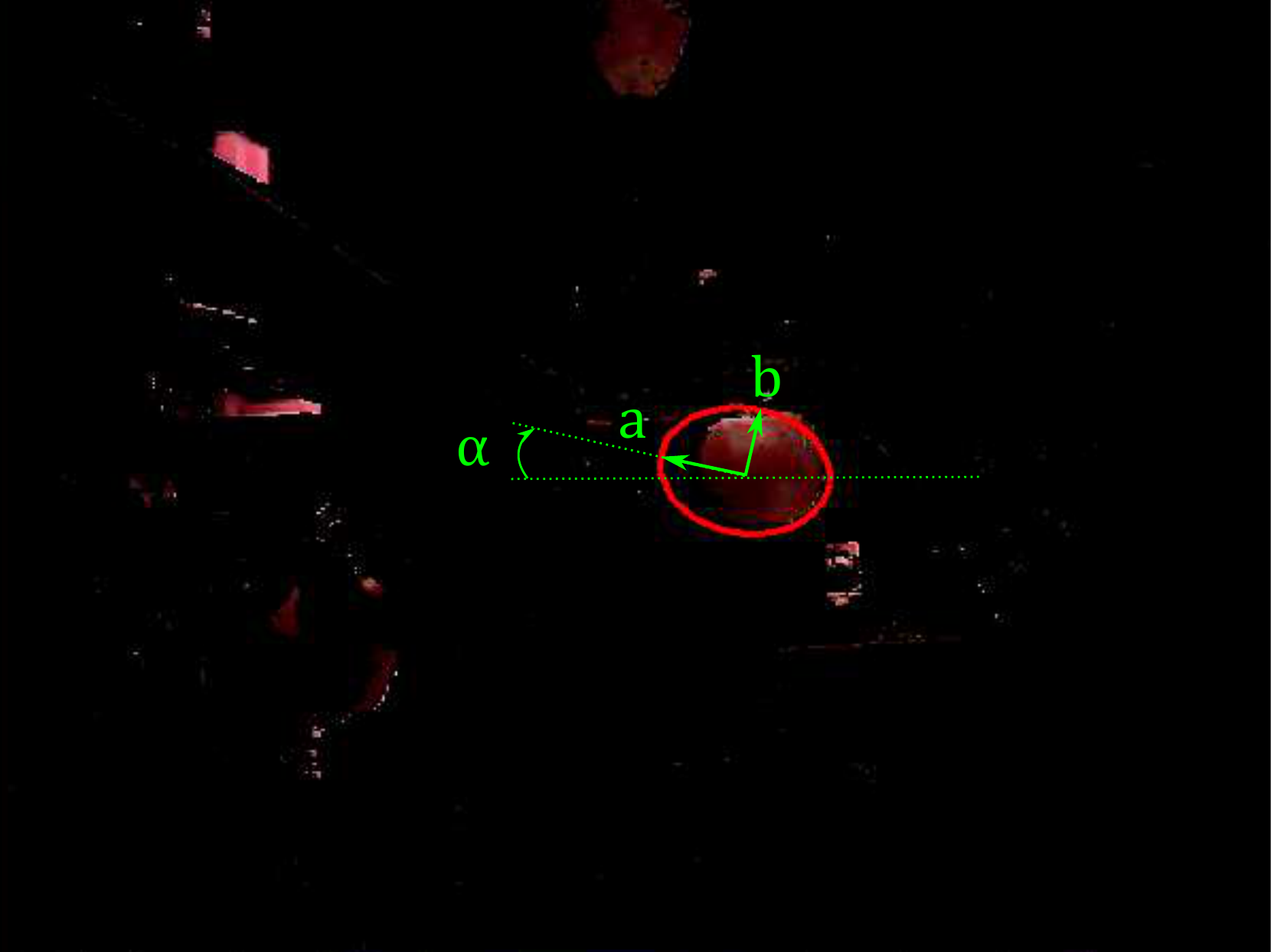}}
\caption{Ellipse tracking using MET in two consecutive images. The contour jumps between two consecutive frames. \label{fig:ellipse}}
\end{figure}

These deformations impair the estimation of the ellipse parameters and lead to ellipses which rotate around their optical center as seen in Figures \ref{fig:ellipse1} and \ref{fig:ellipse2}. This rotational movement introduces large rotations in the computed velocity vector $v_c$ which in turn leads to an oscillating behavior of the robot until the features leave the cameras field of view.

Instead of using manually initialized elliptical features and the MET approach we impose two hypothesis on our feature vector, namely that the ellipse's major axis $a$ and its minor axis $b$ relate roughly as $a \approx b$ and that the rotation angle $\alpha$ between the images $x-$axis and the ellipse's major axis $a$ is $\alpha = 0$. This means that the rotation of the camera around its optical axis becomes obsolete which reduces our controller design from a 6 DOF to a 5 DOF controller.

%% file: initial.tex
\section{Initialization and Visual Servo Controller}
\label{sec:initial}
In Section~\ref{sec:approach} we formulated the assumption that a preprocessing step generates a model of the environment. Further the model of an apple is approximated by a sphere in $\mathbb{R}^3$ which is defined as

\begin{equation}
(X - X_0)^2 + (Y - Y_0)^2 + (Z - Z_0)^2 - R^2 = 0
\end{equation}

with point $(X_0, Y_0, Z_0)$ being the origin of the sphere. To simplify notation we write sphere $S(O, R)$ with $O$ being the spherical models center point and $R$ being the radius of $S$. We use this spherical model to define an initial view plane that is used as the starting point of our visual servoing control law.
Given a point $P_i \in \mathbb{R}^3$ with coordinates $(X_i, Y_i, Z_i)$ which has to be inspected, we define a tangential plane $U$ to point $P_i$ with

\begin{equation}
AX + BY + CZ - D = 0 
\end{equation}

that has normal vector $n = (A, B, C)$. Without computing the plane parameters we can define the plane's normal vector using the sphere center $O$ and point $P_i$ as

\begin{equation}
n = \dfrac{\overrightarrow{O P_i}}{\| \overrightarrow{O P_i} \|} 
\end{equation} 

Using this normal vector we define an initial position of the camera along the normal at distance $d$. In case of availability of a perfect model of the environment and a static target apple we could define distance $d$ between target apple and the camera position to be the final inspection distance. As this is not the case we select distance $d$ big enough so that the whole sphere is guaranteed to be observable considering the dynamics of the system and potential errors in the model.
In Section~\ref{sec:exp:indoor}, we present experiments to identify the region within which the proposed feature selection and tracking approach are robust to deviations from the optimal viewing angle.

We now present the servo law:
A visual servoing approach is defined by minimizing an error function $e(t)$, defined as the difference between an observed visual feature vector $s$ and their desired location $s^*$ in the image plane which can be written as

\begin{equation}
e(t) = s - s^*
\end{equation}

The time variation of the features in the image plane $\dot{s}$ is related to the cameras velocity $v_c$ by

\begin{equation}
\dot{s} = L_sv_c
\end{equation}

in which $L_s$ is called the interaction matrix or the image feature Jacobian, related to $s$. The time variation of the error function is related to the camera velocities similarly by

\begin{equation}
\dot{e} = L_ev_c, \text{where } L_e = L_s
\label{equ:Error}
\end{equation}

Using an exponential decrease of the feature error during the visual servoing is defined as $\dot{e} = -\lambda e$. By using (\ref{equ:Error}) and the exponential error decrease we get an equation that relates the feature error to the cameras velocity by 

\begin{equation}
v_c = -\lambda L_e^{-1} e
\end{equation}

We approximate our target object by a sphere which projects to the image plane as an ellipse with center $(x_0, y_0)$ of the form

\begin{equation}
\begin{split}
&\dfrac{((x-x_0)cos(\alpha)+ (y-y_0)sin(\alpha))^2}{a^2} \\ 
&+ \dfrac{((x-x_0)sin(\alpha)+(y-y_0)cos(\alpha))^2}{b^2} -1 = 0
\label{equ:Ellipse}
\end{split}
\end{equation}

This ellipse is parametrized by 5 parameters: its center point $(x_0, y_0)$, major axis $a$,  minor axis $b$ and the angle $\alpha$ between the major axis and the image's $x-$axis. A visual servoing approach using ellipse features was proposed first by Chaumette~\cite{chaumette_image_2004}. Instead of using the angular function $\alpha$ directly he proposed to use parameter $t = tan(\alpha)$.

It can easily be seen that this approach becomes degenerate in the case when the projection of the sphere becomes a circle as $t$ is then undefined. This is of course always the case when the cameras optical axis and the sphere center are aligned which would make this approach unsuitable for our purpose. Chaumette~\cite{chaumette_image_2004} addressed this problem by using the ellipse's center of gravity $(x_0, y_0)$ and the normalized inertial moments $(\mu_{20}, \mu_{11}, \mu_{02})$ to parametrize the feature ellipses. In this case $\mu_{20}$ and $\mu_{02}$ describe the variance in $x$ and $y$ direction respectively and the moment $\mu_{11}$ is the covariance between $\mu_{20}$ and $\mu_{02}$ and describes the orientation of our circle. By defining the observed feature vector as $F_o = s(x_0, y_0, \mu_{02}, \mu_{11}, \mu_{20})$ and the target features as $F_d = s^*(x_0^*, y_0^*, \mu_{02}^*, \mu_{11}^*, \mu_{20}^*)$ we compute the visual servoing error $e = s - s^*$ as 

\begin{equation}
e = s - s^* = \begin{bmatrix}
x - x^* \\
y - y^* \\
\mu_{20} - \mu_{20}^* \\
\mu_{11} - \mu_{11}^* \\
\mu_{02} - \mu_{02}^*
\end{bmatrix}
\label{equ:errorFeatures}
\end{equation}

Using this feature vectors leads to the interaction matrix whose derivation is presented in~\cite{chaumette_image_2004}. We note that the interaction matrix requires computation of the feature depth $Z$ which is not trivial. In our case we can use the results of the preprocessing step which gives us an estimate of the model and it's parameters. This way, we obtain  an approximation of the target apple's diameter $d_A$ with which we compute the feature depth $Z$ as follows

\begin{equation}
Z = -f \dfrac{d_A}{w_p \mu_p}
\end{equation}

with the cameras focal length $f$ (mm), the observed ellipses diameter $w_p$ (pixel), the image pixel size $\mu_p$ (mm/pixel) and the model's diameter $d_A$ (mm). Lastly we define our convergence criteria for the algorithm  as the sum of squared feature errors which can be written as

\begin{equation}
e = \sum_{i=1}^n (s_i - s_i^*)^2 
\end{equation}

%% file: vision.tex
\section{THE VISUAL FEATURE TRACKER}
\label{sec:controller}
Tracking methods such as the moving edges tracker \cite{marchand_feature_2005} have been shown to work well in indoor environments.
Testing these methods in the apple orchard setting revealed three challenges when using such an out of the box tracker:

\begin{itemize}
\item The target has to be specified manually.
\item The tracker is not robust to illumination changes.
\item In a dynamic outdoor environment the standard tracker loses track of the manually specified target apple after an average of three iterations.
\end{itemize}

\begin{figure}[ht!]
\vspace{0.2cm}
	\centering
	\includegraphics[width=0.9\columnwidth]{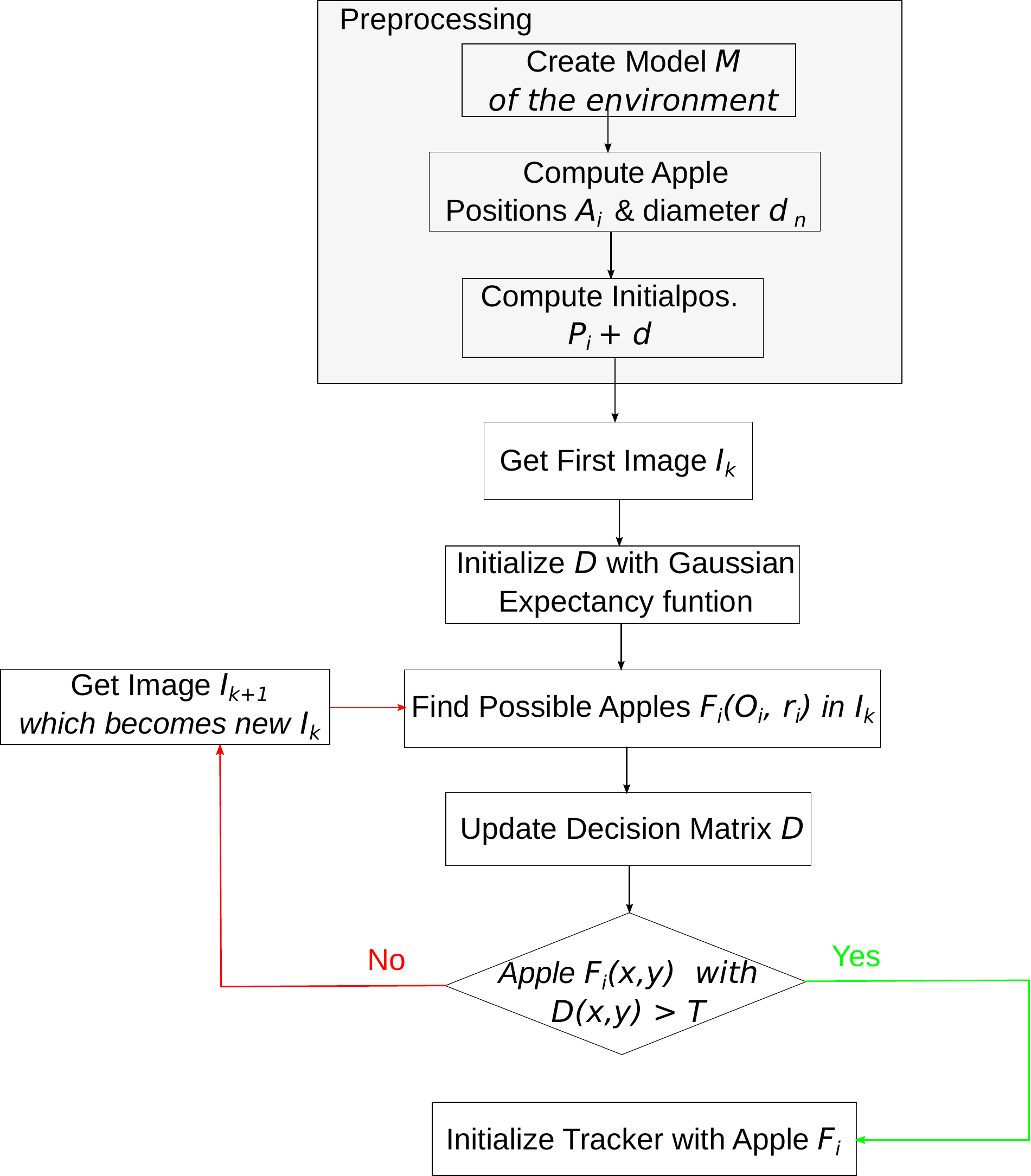}
	\caption{Preprocessing and initialization of the feature tracker}
	\label{fig:processing}
\end{figure}

To address these challenges, we developed a custom visual feature detection and tracking approach which is initialized as seen in Figure \ref{fig:processing}.

A first image $I_k$ is acquired in which we search for possible target apples $F_i(O_i, r_i)$ where $O_i(x_i, y_i)$ is the observed projections center point. This search is carried out by an adaptive Hough transform (HT) search in the image. Using only a Hough transform as tracking input would have various shortcomings:
(i)~The HT depends on various parameters that have to be tuned for each image which makes it unsuitable for a fully automatic approach.
(ii)~The HT does not provide a mechanism for accurate selection of a target out of multiple possibilities according to spatial location and radii criteria.
(iii)~If the parameters do not match well, the HT algorithm returns possible targets that are not apples. 
For our adaptive HT we vary the parameters from coarse to fine to find an unsorted structure of possible targets and use this only as an estimate of possible targets. This is a necessary step as an image may very well include more than one apple at any time. 
To select a target apple from the set of possible targets we create a decision matrix $D$. This matrix is initialized with a kernel $K$ which maps our expectation of location and radius of the projection of our selected apple in the image plane. In other words we use a Gaussian kernel with expected projection radius $r_p$ at the image's principal point $O_p(x_p, y_p)$.

\begin{figure}[h!]
\centering     
\subfigure[Initialized decision matrix]{\label{fig:initial}\includegraphics[width=0.45\columnwidth]{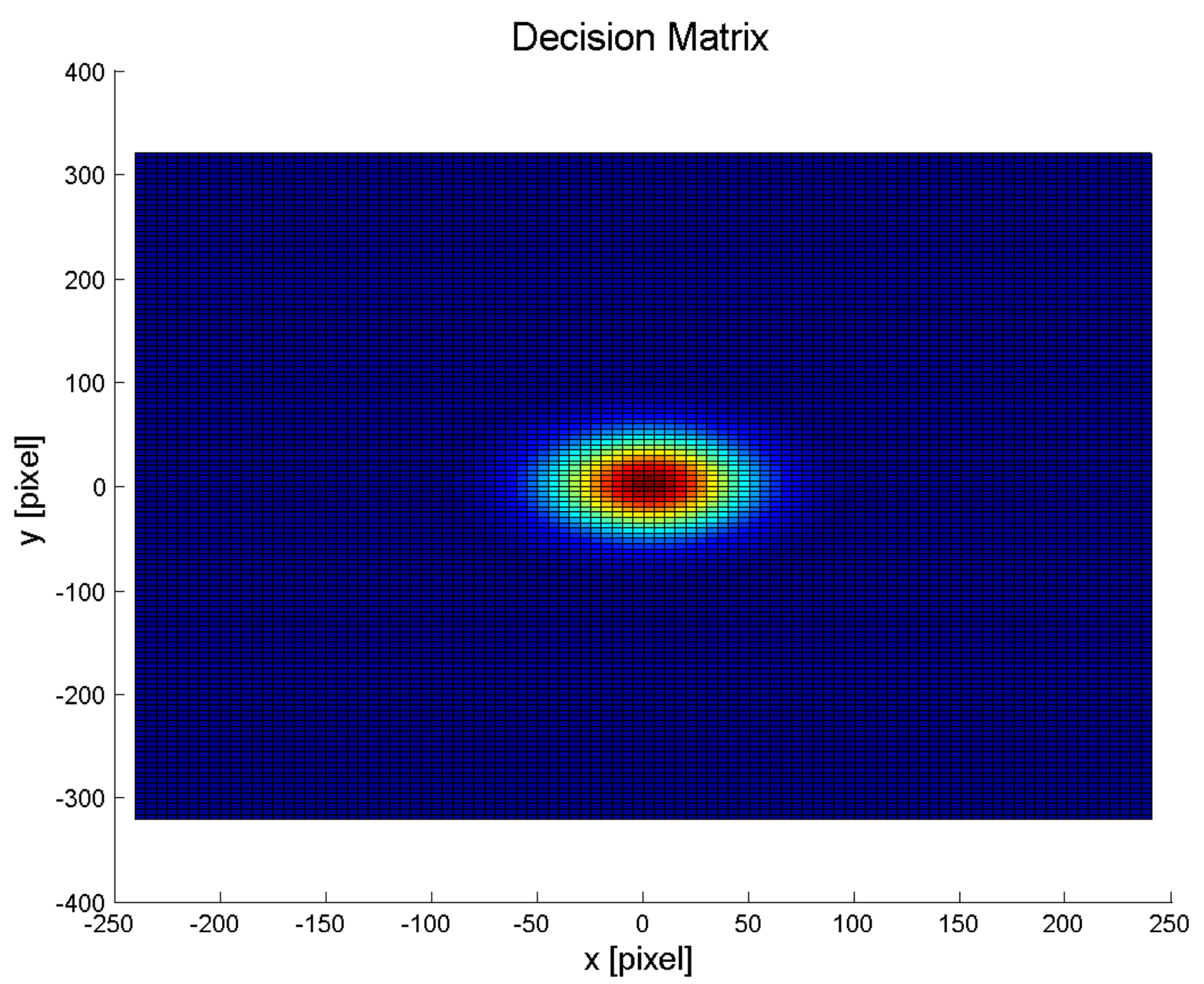}}
\subfigure[Updated decision matrix]{\label{fig:updated}\includegraphics[width=0.45\columnwidth]{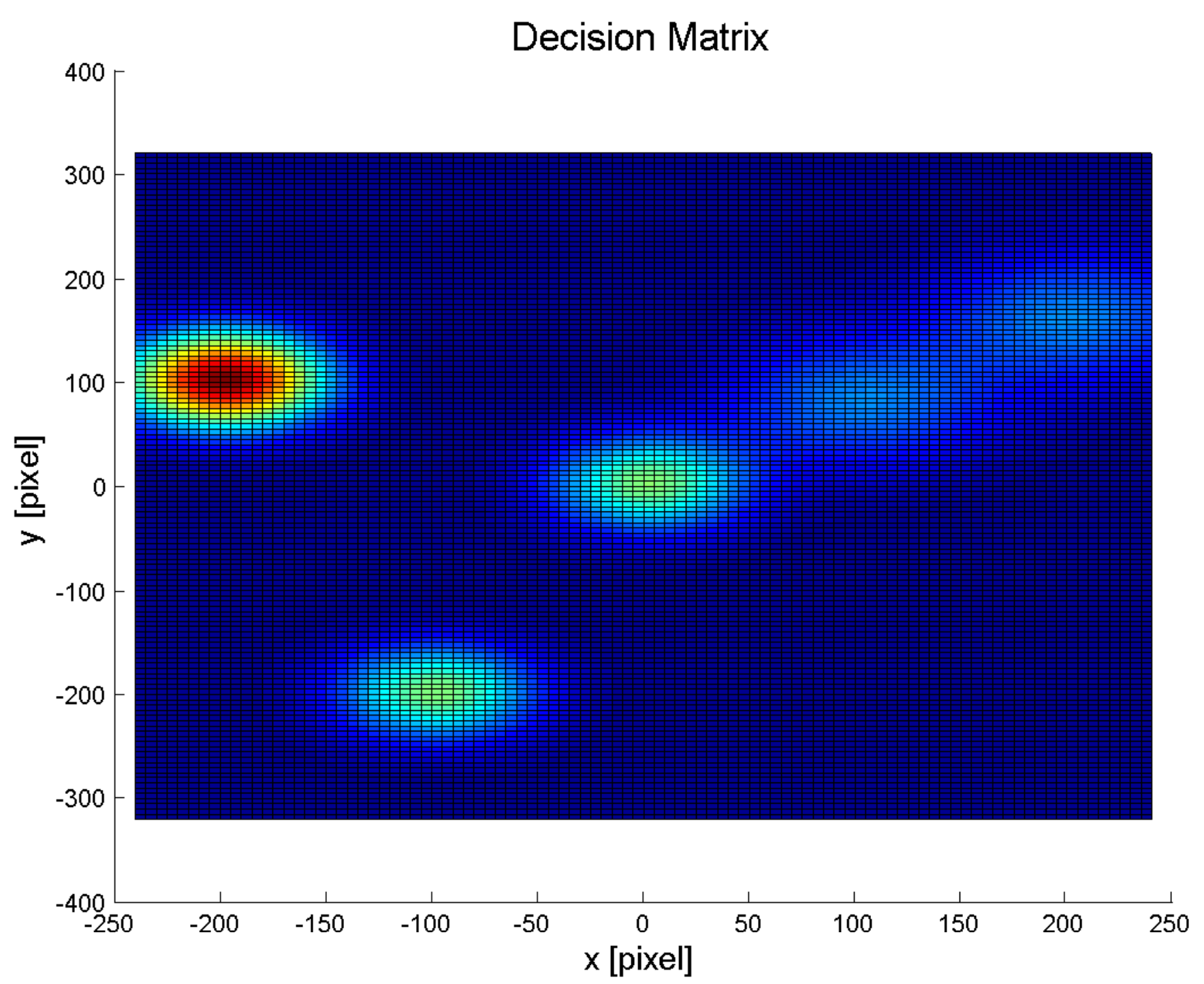}}
\caption{Decision Matrix initialized and after update cycles}
\end{figure}

After initialization of the decision matrix we use the found vector of possible targets  to update the decision matrix using weighted probabilities for each observed candidate target to create kernels $K_i$. In a first step we sort all target apples according to a confidence score $c_s$ that takes the spatial location and the projected radius into consideration as

\begin{equation}
c_s = \left| \dfrac{x - x_0}{x_0} \right| + \left| \dfrac{y - y_0}{y_0} \right| + \left| \dfrac{r - r_0}{r_0} \right|
\end{equation}

with $(x, y, r)$ being the center point and radius of the observed feature projection and $(x_0, y_0, r_0)$ being the ideal center point and radius of the projection in the image. We use our confidence score to sort the probable targets.

In the next step we assign evenly distributed, normalized weights to the sorted targets $F_{iS}$. If the adaptive HT returns $i = (1, .., N)$ possible targets then we assign a weight $w_i$ to the sorted targets with 

\begin{equation}
w_i = \dfrac{i}{\sum_{i=1}^N i}
\end{equation}
 
From our initialization process we know that a target apple is present in the image space. The probability that the desired apple is contained in the image is therefore

\begin{equation}
P(I | F_s) = 1 
\end{equation}

From this we are able to write the probability assigned to each possible target as

\begin{equation}
P(I | F_{iS}) = P_i(I | F_s) w_i
\end{equation}

Using these probabilities we define updating kernels $K_i$ as

\begin{equation}
K_i = w_i^{\left( \dfrac{(x- x_i)^2}{2r^2} + \dfrac{(y-y_i)^2}{2r^2} \right)}
\end{equation}

$K_i$ is another Gaussian kernel that is computed using the possible targets center point, the observed features radius $r$ which is the average of the minor and major axis of the observed ellipse and the amplitude $V = P(I | F_{iS})$. The decision matrix is updated with each kernel by adding them to the decision matrix at their prospective location $(x_i, y_i)$. The usage of position and radius to compute the sorting of our weighting scheme for potential apples increases the robustness of the whole tracking approach as it filters out false positives. This update process is repeated for every image $I_k \gets I_{k+1}$ until a kernel meats the selection criteria $S_c$.

Our selection criteria is defined as $S_c = D(x,y) > T$ where $T$ is a threshold that is empirically selected to require a minimum of two and a maximum of five update cycles to find our selected target apple. Next the visual feature tracker is initialized with the current image and the decision matrix.

Using this target we define a circular ROI $R$ that contains the target which becomes our new search space. Using the freeman chain code in this region we find a vector of $n$ boundaries from which we use the longest one as the representation of the boundary of our target. Using the boundary points $p_i$ we compute an elliptical model for the observed target, using the RANSAC algorithm to eliminate boundary points that do not belong to the target's boundary such as defects or holes.

\begin{algorithm}[ht!]
\begin{algorithmic}[1]
  \REQUIRE $I_k$, Selected Apple features $F_s$ from initialization (Figure~\ref{fig:processing})
  \STATE Update $D$ with the position of the selected $F_s(x_s, y_s, r_s)$
  \STATE Create Region Of Interest (ROI) $R(x_s, y_s,r_s)$ from $F_s$ that satisfies $S_c$
  \WHILE {$F_s(x_s, y_s) != I_k(x_p, y_p)$}
  \STATE $I_{k} \gets I_{k+1}$
  \IF{$F_s$ not found in $R$ and $R < I_k$}
  \STATE Increase size of $R$
  \ELSIF{$R > I_k$}
  \STATE Reset $D$ to force reinitialization of the tracker  
  \ELSE
  \STATE Update $D$ and $R$ with $F_s(x_s, y_s, r_s)$
  \STATE Compute feature vector $F_o(x_o, y_o, \mu_{02}, \mu_{11}, \mu_{20})$ from boundary points $p_i(x_i, y_i)$
  \STATE Pass $F_o$ to the visual servoing control law
  \ENDIF
  \ENDWHILE
\end{algorithmic}
\label{alg:featureTracker}
\caption{Feature tracking algorithm after initialization}
\end{algorithm}

Using the left over points we compute the observed feature vector $F_o(x_o, y_o, \mu_{02}, \mu_{11}, \mu_{20})$ with $(x_0, y_0) = \dfrac{1}{N} \sum_i^N x_i, \dfrac{1}{N}\sum_i^N y_i$ and the moments are computed using
\begin{equation}
\mu_{ij} = \sum_x\sum_y(x-x_0)^i(y-y_0)^i
\end{equation}

The robustly estimated feature vector $F_o$ is the tracker's output and is passed to the visual servoing algorithm. Using our tracking approach guarantees the robustness of the visual servoing controller with respect to occlusion and illumination changes. 
To update the tracker we repeat the steps until convergence is achieved. This assures that the once selected apple is being tracked as we continuously us information from former frames $I_{k-1}$. This is embedded in the decision matrix. One can see that if the tracker does not find a valid target in $R$ we increase the size of the ROI incrementally to the whole image space. If the ROI consists of the whole image space we reset $D$ and force a reinitialization of the tracker.

%% file: expt.tex
\section{EXPERIMENTS}
\label{sec:exp}
For the validation of our approach we extensively tested the system in indoor and outdoor settings. 
The reader can view videos of some of these experiments in the accompanying video.

\subsection{Indoor Experiments} \label{sec:exp:indoor}
To evaluate the initial behavior of the system, we first performed experiments in a controlled, indoor setup. Controlled refers here to constant illumination and static target objects as there is no wind. For the indoor experiments a 3D model of an artificial tree was constructed from which the apple locations and initial sensor position were computed. The initial position was selected $0.5\mathrm{m}$ away from the apple in $-y-$direction and as a convergence metric we use the residual sum of squared feature errors which has to converge to $e < 0.0001 mm^2$ between desired and observed features.

For the following experiments we show three Figures:
(i)~The first figure shows the target apple at the initial position. The target apple's observed projection is shown in green and the desired projection is shown in red. 
(ii)~The second figure shows the target apple when convergence is achieved.
(iii)~Finally we plot the error convergence and the computed velocity vectors. 

\begin{figure}[h!]
\centering     
\subfigure[Initial position]{\label{fig:a}\includegraphics[width=0.45\columnwidth]{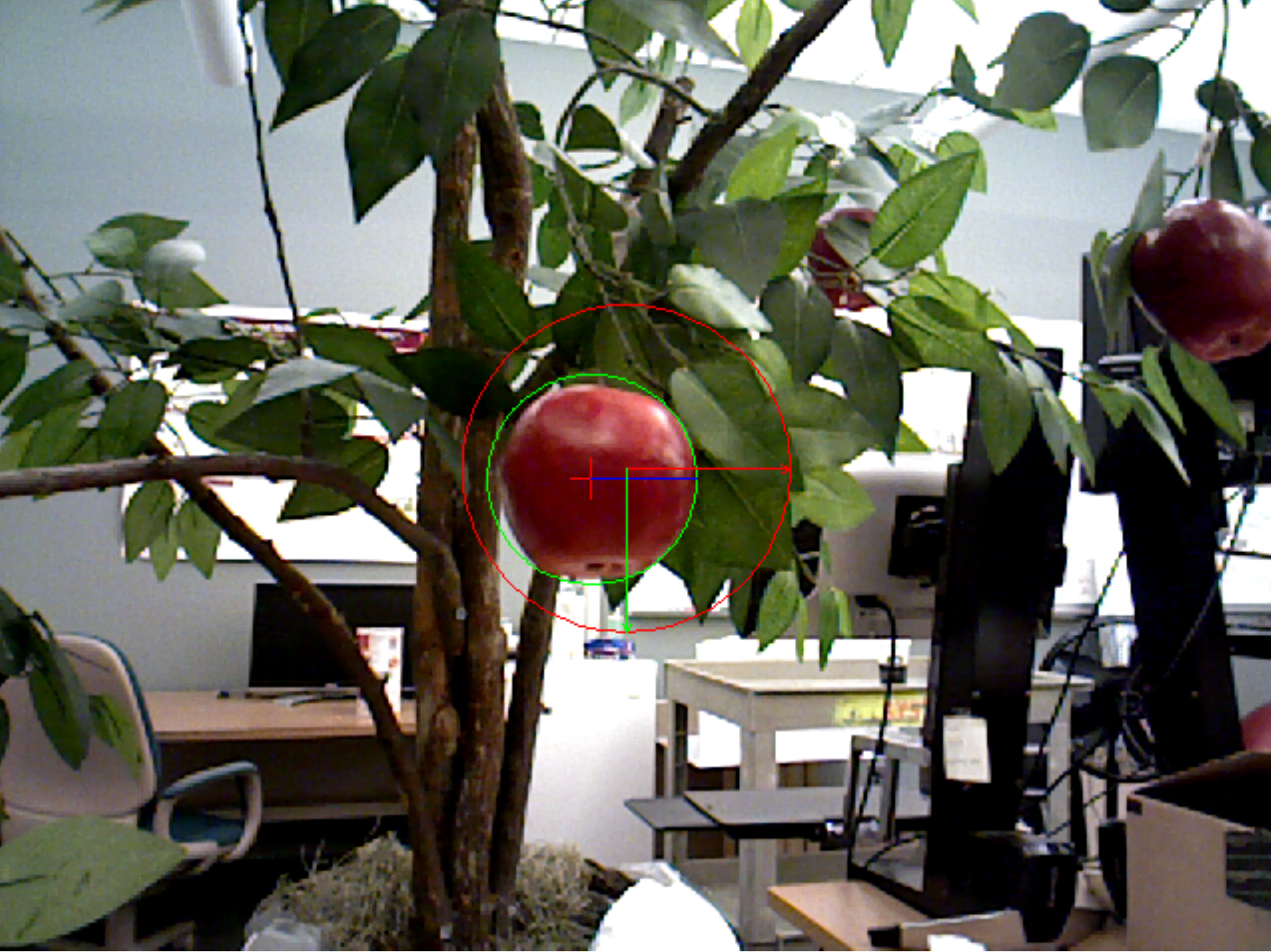}}
\subfigure[Converged position]{\label{fig:b}\includegraphics[width=0.45\columnwidth]{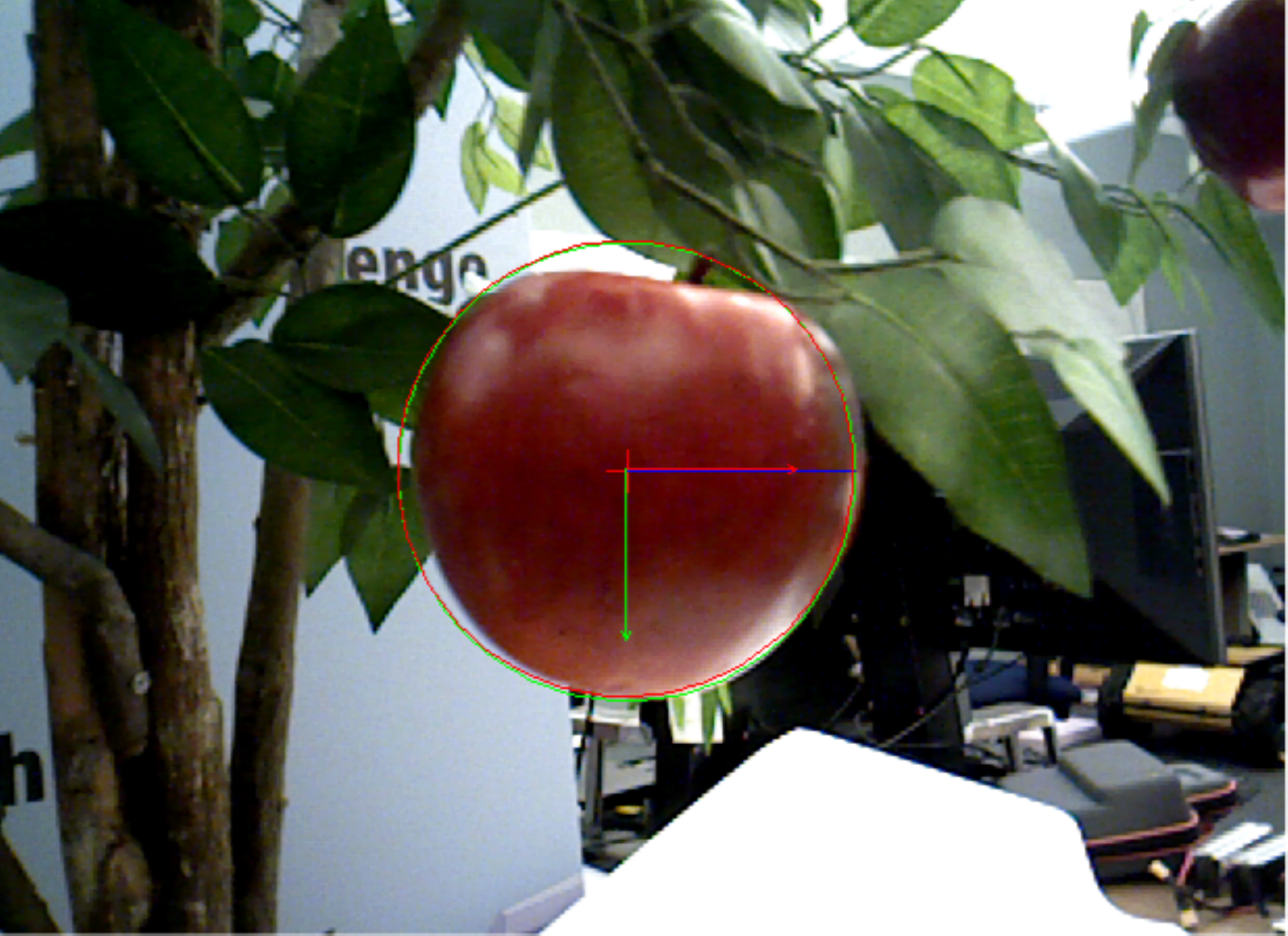}}

\subfigure[Velocity and error convergence]{\label{fig:c}\includegraphics[width=\columnwidth]{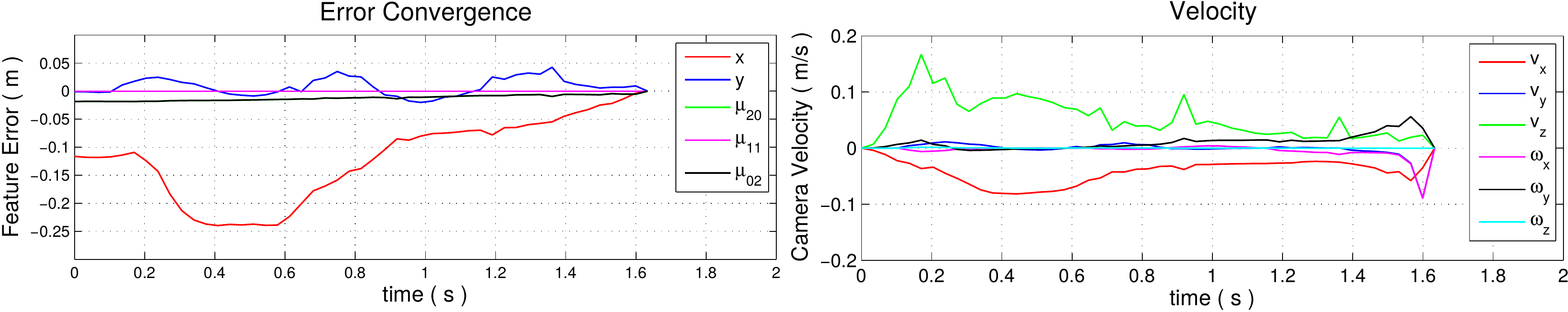}}
\caption{Indoor experiment under controlled conditions \label{fig:Indoor}}
\end{figure}

For the first experiment we observe that the apple is initially in proximity of the principal point of the image. The error in $y-$direction first gets larger as the robot tries to minimize the maximal error which is in the cameras $z-$axis. After about $1.5\mathrm{s}$ the error converges.

\begin{figure}[h!]
\centering     
\subfigure[Initial position]{\label{fig:deseaseA}\includegraphics[width=0.45\columnwidth]{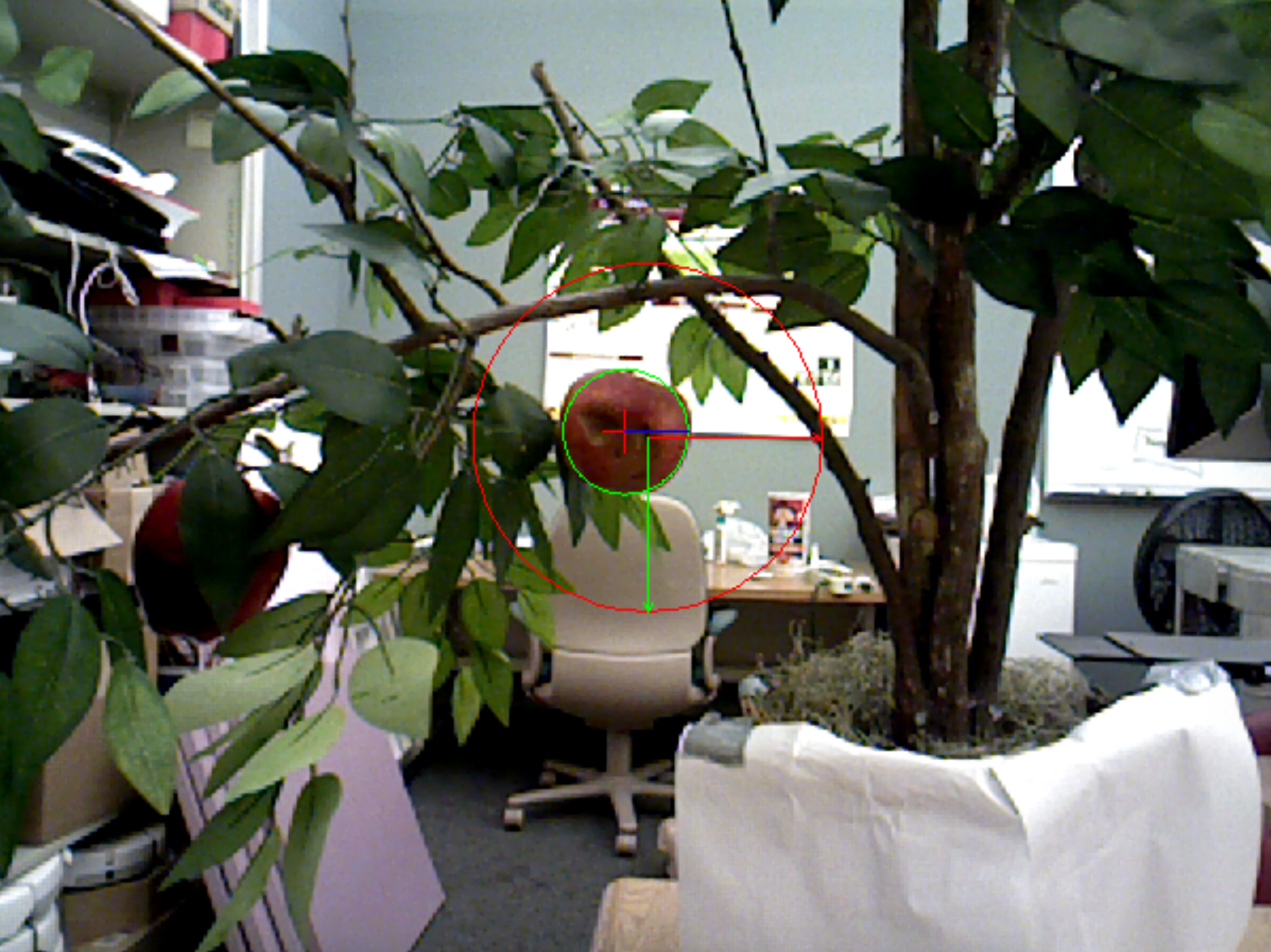}}
\subfigure[Converged position]{\label{fig:deseaseB}\includegraphics[width=0.45\columnwidth]{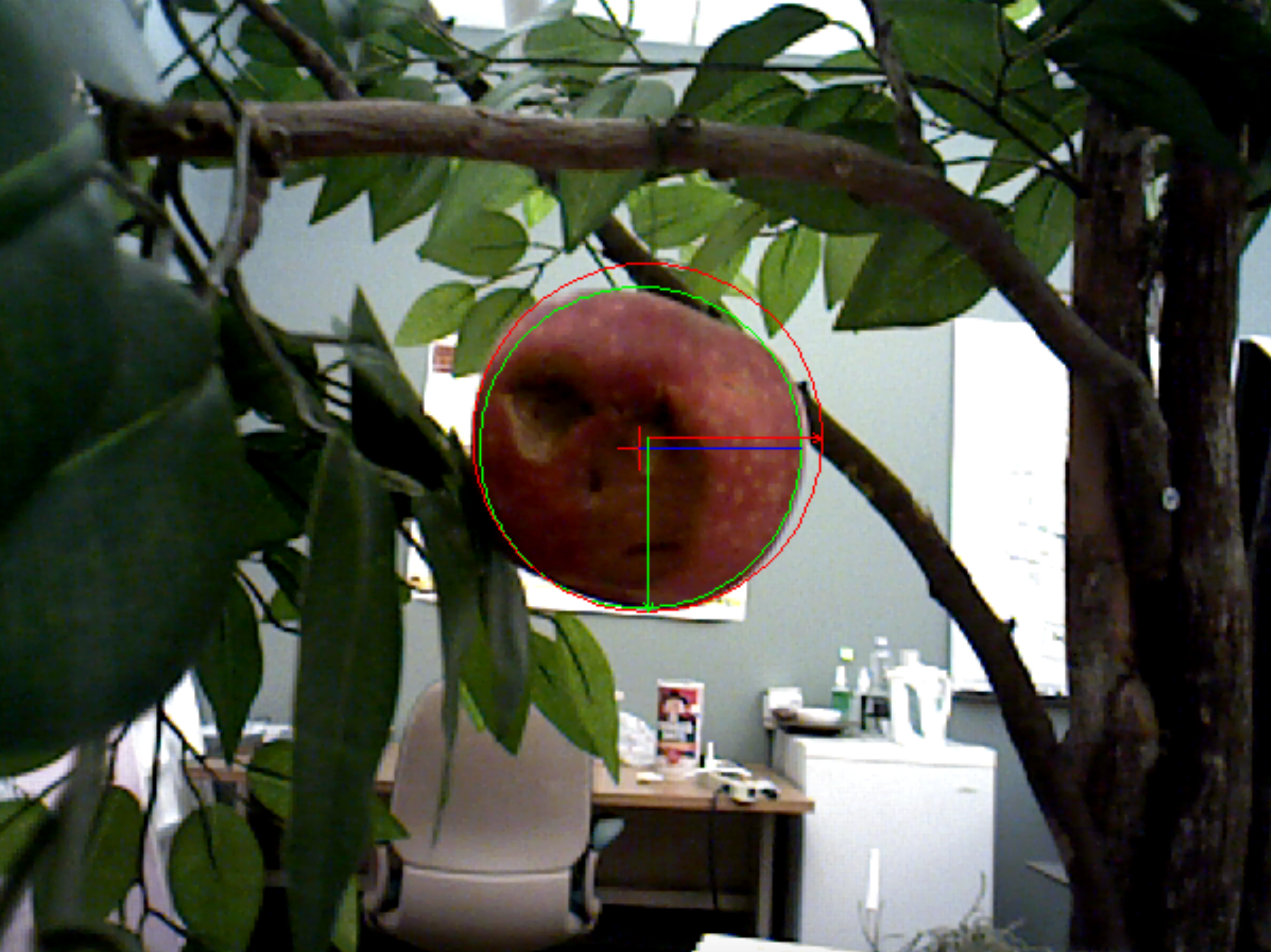}}

\subfigure[Velocity and error convergence]{\label{fig:deseaseC}\includegraphics[width=\columnwidth]{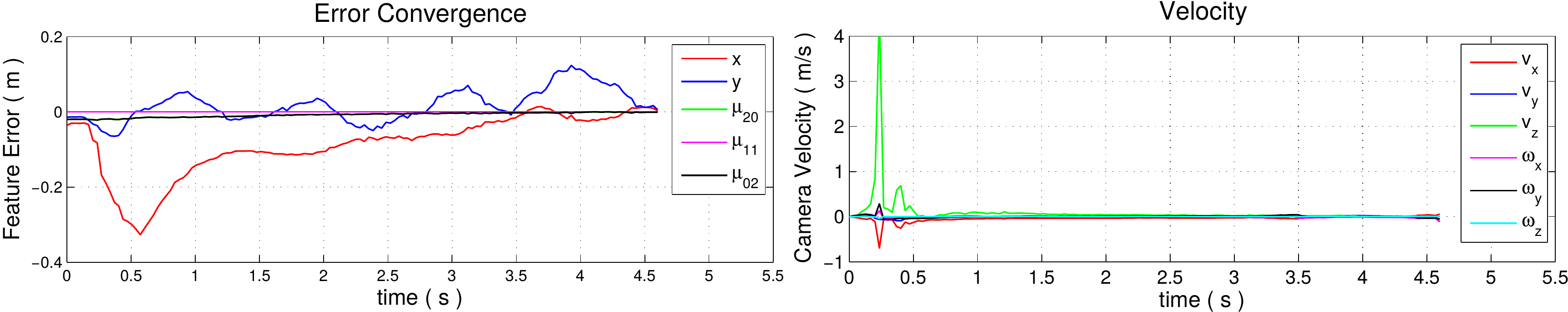}}
\caption{Close up inspection of a real fruit with disease 
\label{fig:disease}}
\end{figure}

In the next experiment the convergence was tested using an apple with diseases which are visually detectable as patches of dark texture and holes in the apples surface. The error in $y-$direction first increases while the error control law tries to minimize the main error in $z-$direction. The time increase is due to the increased distance between initial position and target which makes the robot approach a singularity which leads to slow movements. One notices a large peak in the camera velocity. This is due to a large initial error and due to the fact that the adaptive gain increases the velocity. However we put limit maximal possible robot velocity to feasible values.

\begin{figure}[h!]
\centering     
\subfigure[Initial position]{\label{fig:LowerLeftA}\includegraphics[width=0.45\columnwidth]{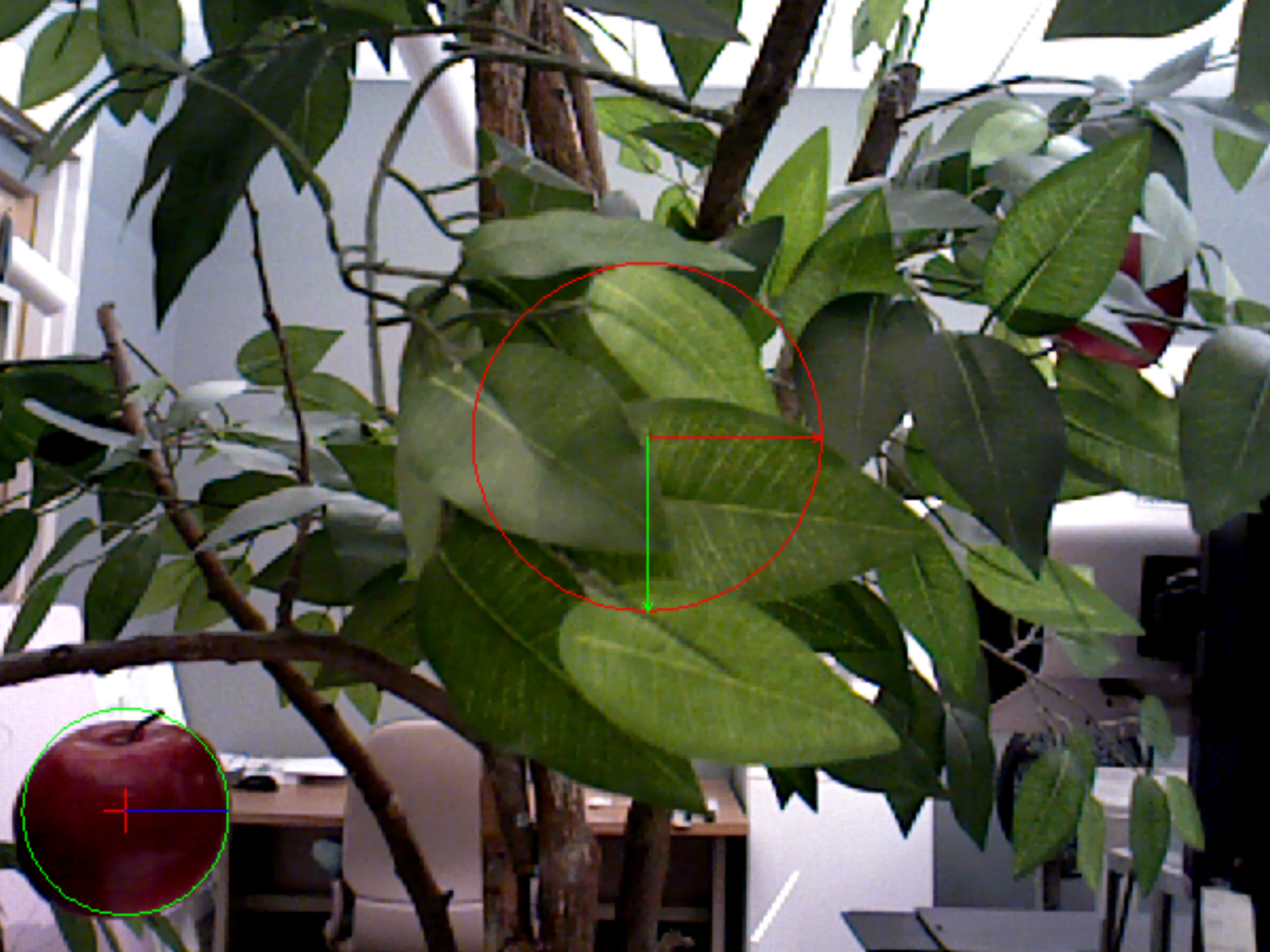}}
\subfigure[Converged position]{\label{fig:LowerLeftB}\includegraphics[width=0.45\columnwidth]{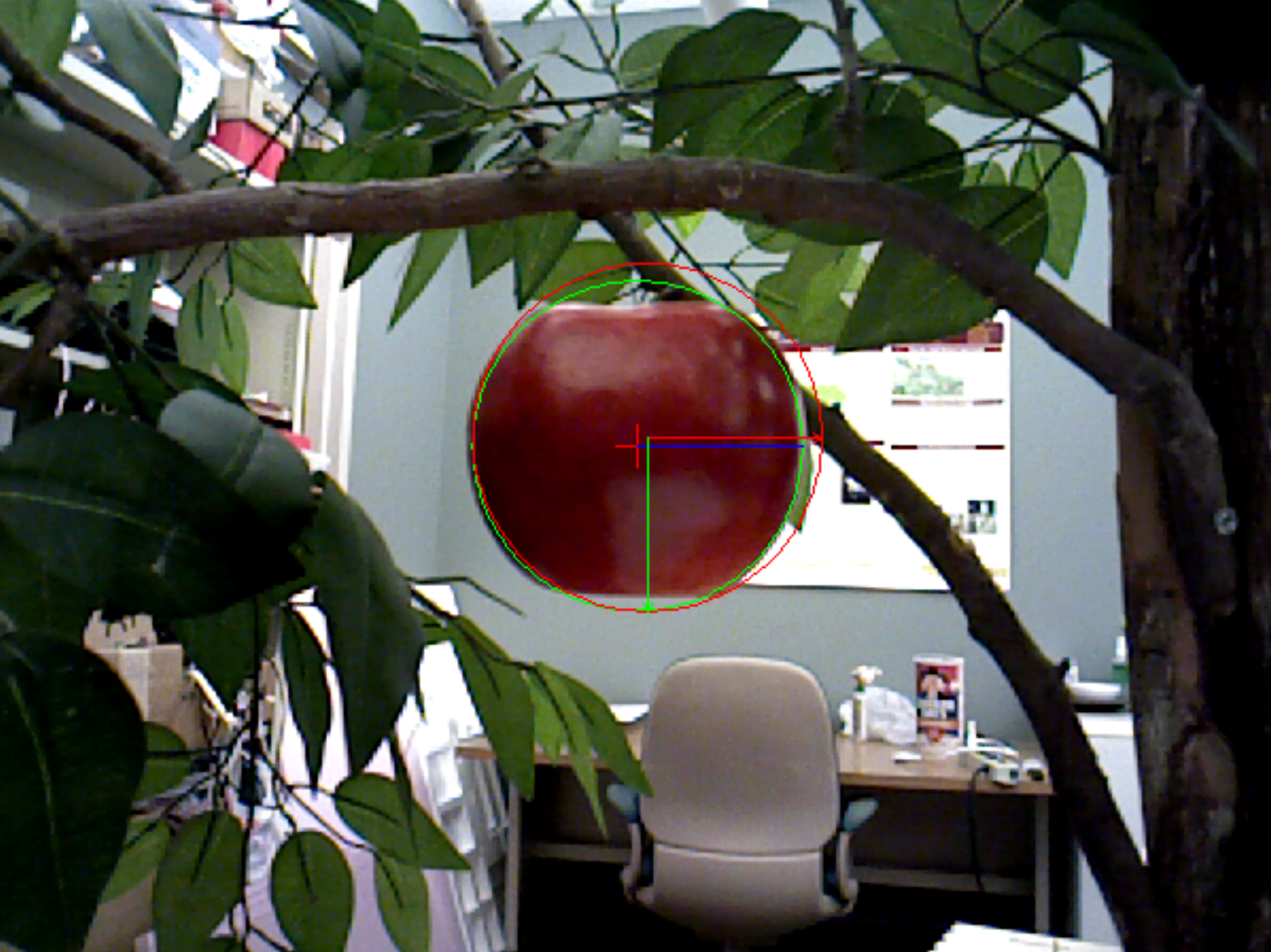}}

\subfigure[Velocity and error convergence]{\label{fig:LowerLeftC}\includegraphics[width=\columnwidth]{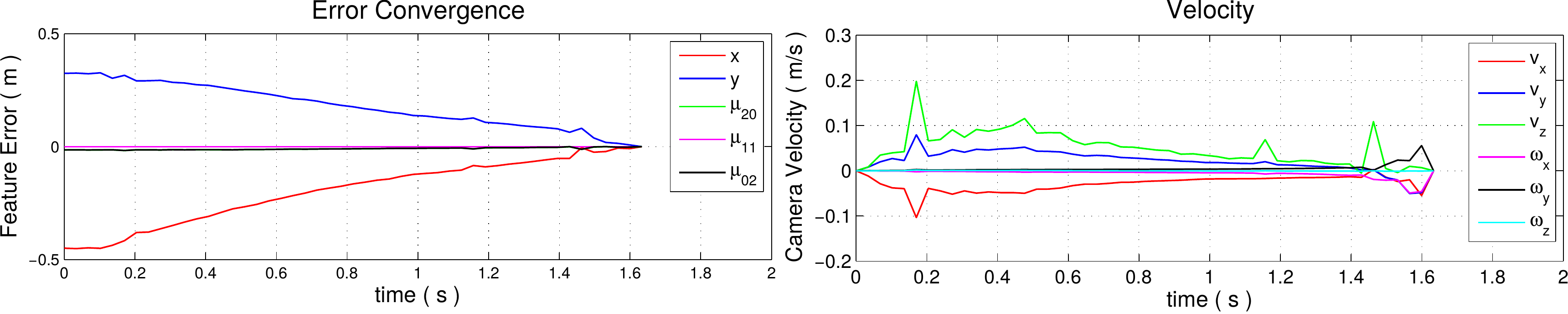}}
\caption{Convergence when target location is at the lower left image corner \label{fig:lowerLeft}}
\end{figure}

In the next experiment it can be observed that the error converges in cases where the target object is present in the lower left or the upper right corner of the image.
The viewing cone for which we can guarantee convergence of the visual feature error is limited by the viewing angle of the sensor which is $43\mathrm{\degree}$ in vertical and $57\mathrm{\degree}$ in horizontal direction and the workspace of the manipulator which is in the interval $[0, 0.8] \mathrm{m}$.

\begin{figure}[h!]
\centering     
\subfigure[Initial position]{\label{fig:UpperRightA}\includegraphics[width=0.45\columnwidth]{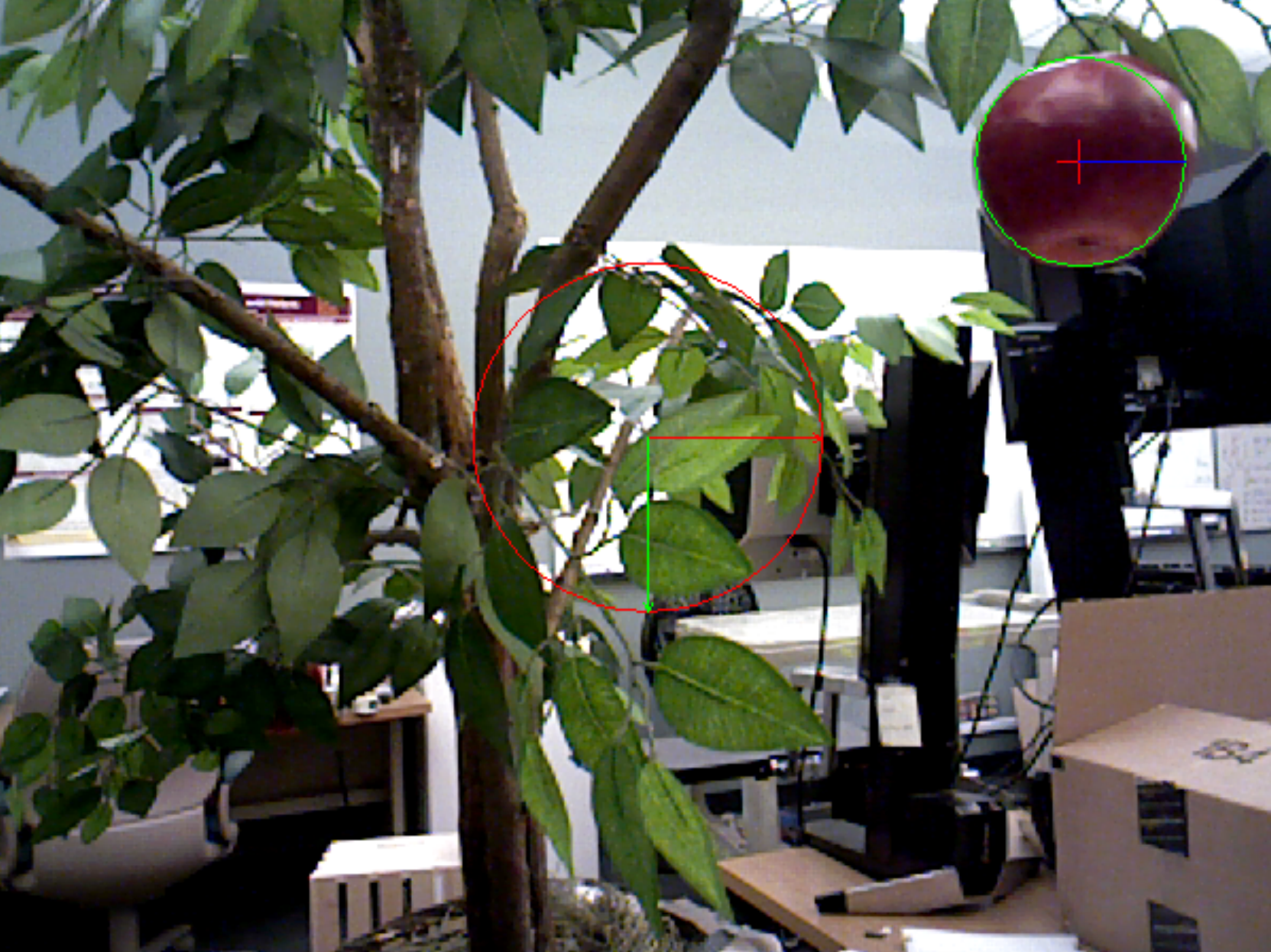}}
\subfigure[Converged position]{\label{fig:UpperRightB}\includegraphics[width=0.45\columnwidth]{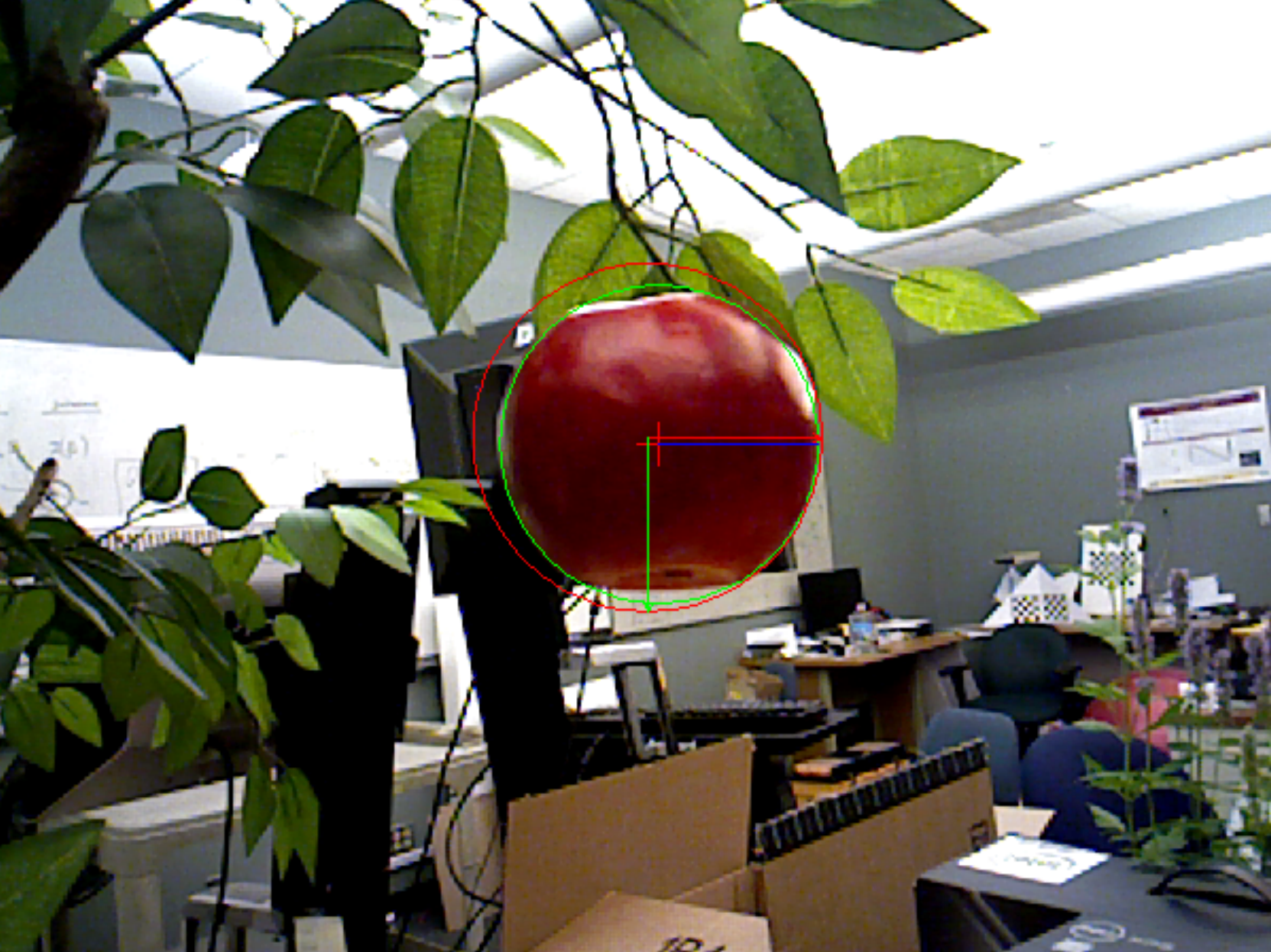}}

\subfigure[Velocity and error convergence]{\label{fig:UpperRightC}\includegraphics[width=\columnwidth]{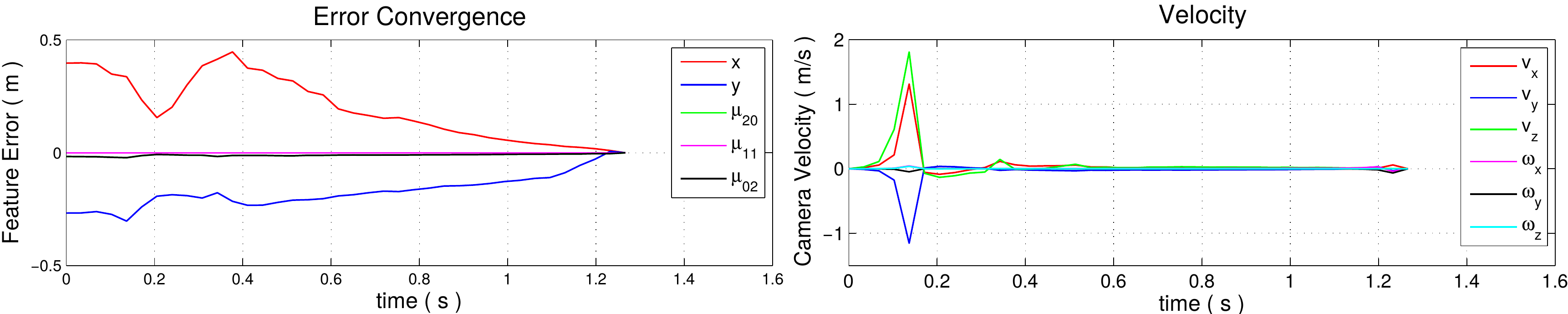}}
\caption{Convergence when target location is in the upper right image corner \label{fig:upperRight}}
\end{figure}

\subsection{Field Experiments}
The field experiments were conducted at the UMN Landscape Arboretum Horticultural Research Center in Eden Prairie, MN. In the outdoor experiments we had no 3D model of the environment and hence no model from which to compute the initial position. To show convergence in outdoor settings we randomly selected a position of the robot approximately $0.5\mathrm{m}$ away from the target in direction of the robot's $-y$ axis. However as these initial positions are only approximations of the true sensor locations and no ground truth is available the convergence times of the outdoor experiments can not be directly compared with each other or the indoor experiments.

\begin{figure}[h!]
\centering     
\subfigure[Initial position]{\label{fig:d}\includegraphics[width=0.45\columnwidth]{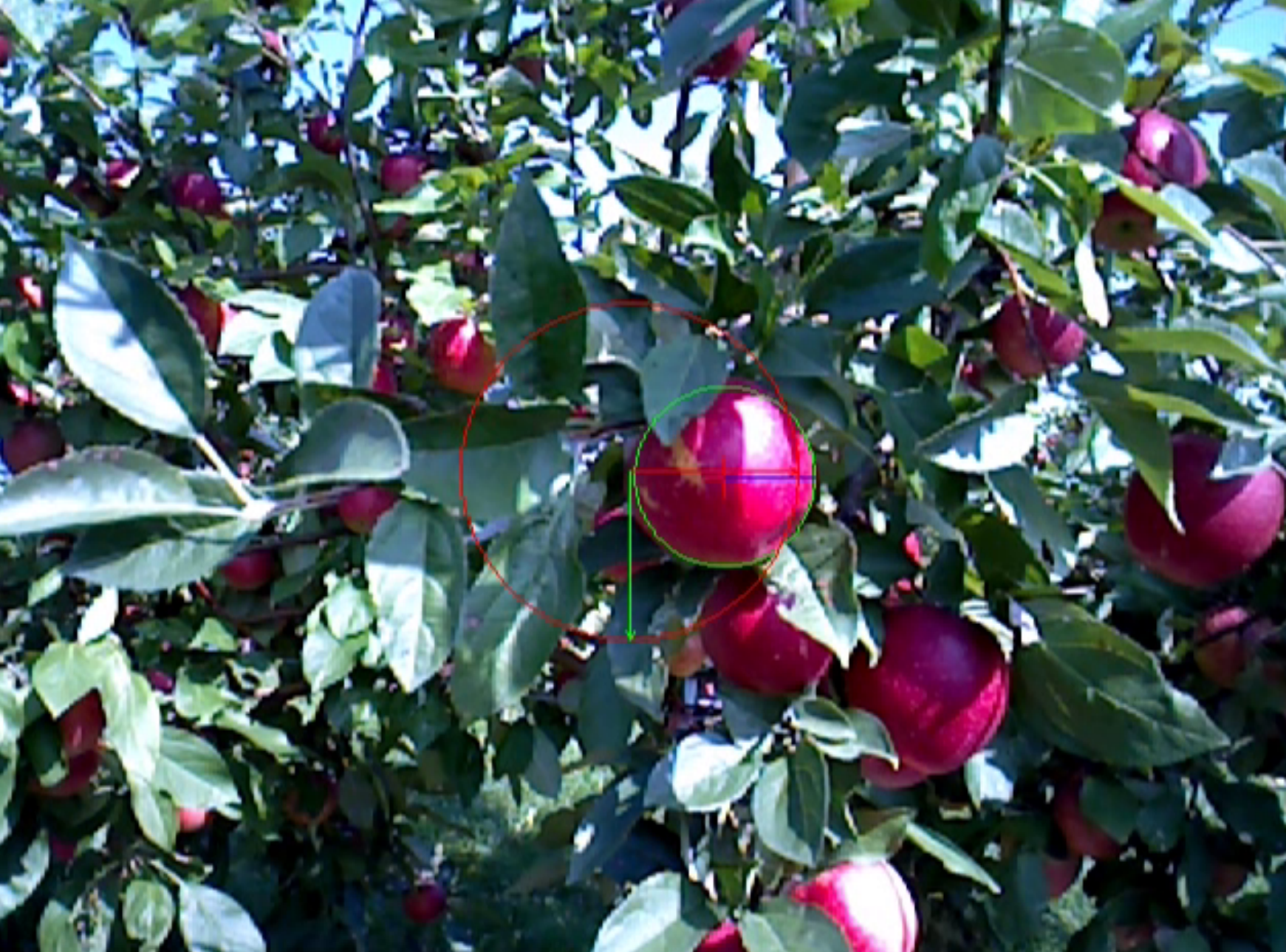}}
\subfigure[Converged position]{\label{fig:e}\includegraphics[width=0.45\columnwidth]{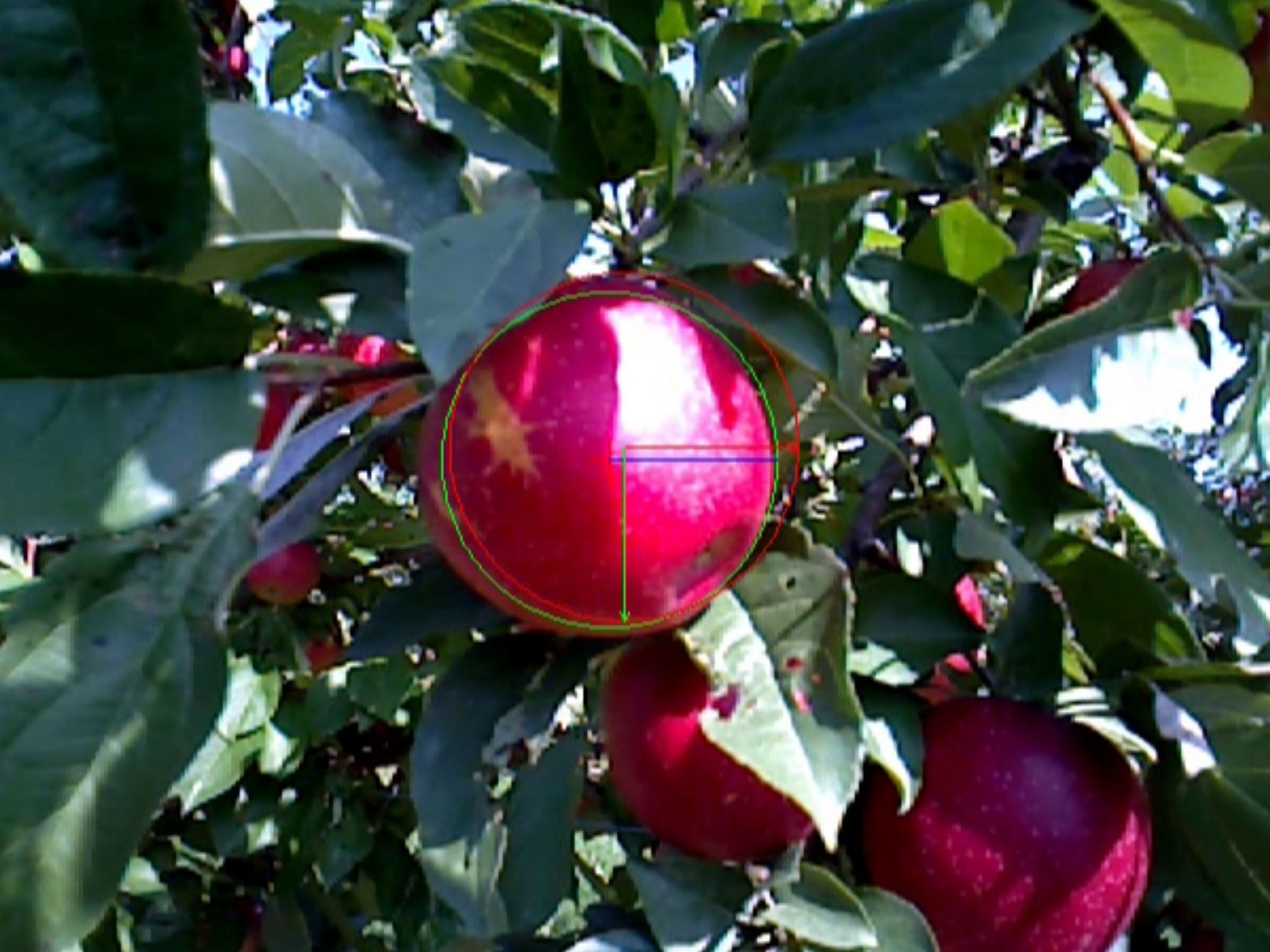}}

\subfigure[Velocity and error convergence]{\label{fig:f}\includegraphics[width=\columnwidth]{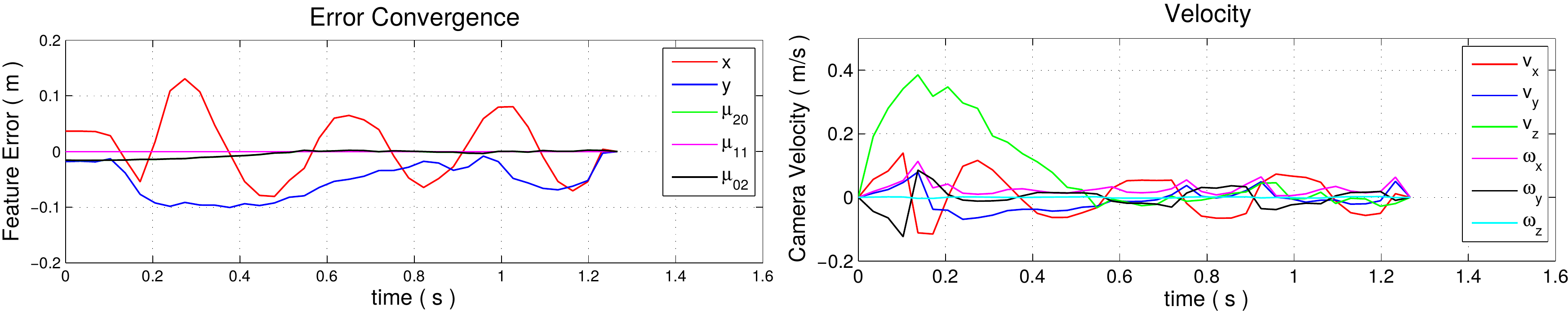}}
\caption{Outdoor experiment with light wind (avg. $16\mathrm{km/h}$) \label{fig:fieldExperiment}}
\end{figure}

The experiments were conducted on a sunny day so that the apple surface is partially in the shadow and shows strong specular reflections on other parts of the apple. Further during the experiment the wind conditions were alternating with average wind speeds of $16\mathrm{km/h}$ and peaks of up to $25.6\mathrm{km/h}$ according to the National Weather Service. In Figure \ref{fig:f} we see the behavior of the control algorithm in such conditions. A slight oscillation of the feature error can be observed which is due to this system dynamics. 

\begin{figure}[h!]
\centering     
\subfigure[Initial position]{\label{fig:buttomUpA}\includegraphics[width=0.45\columnwidth]{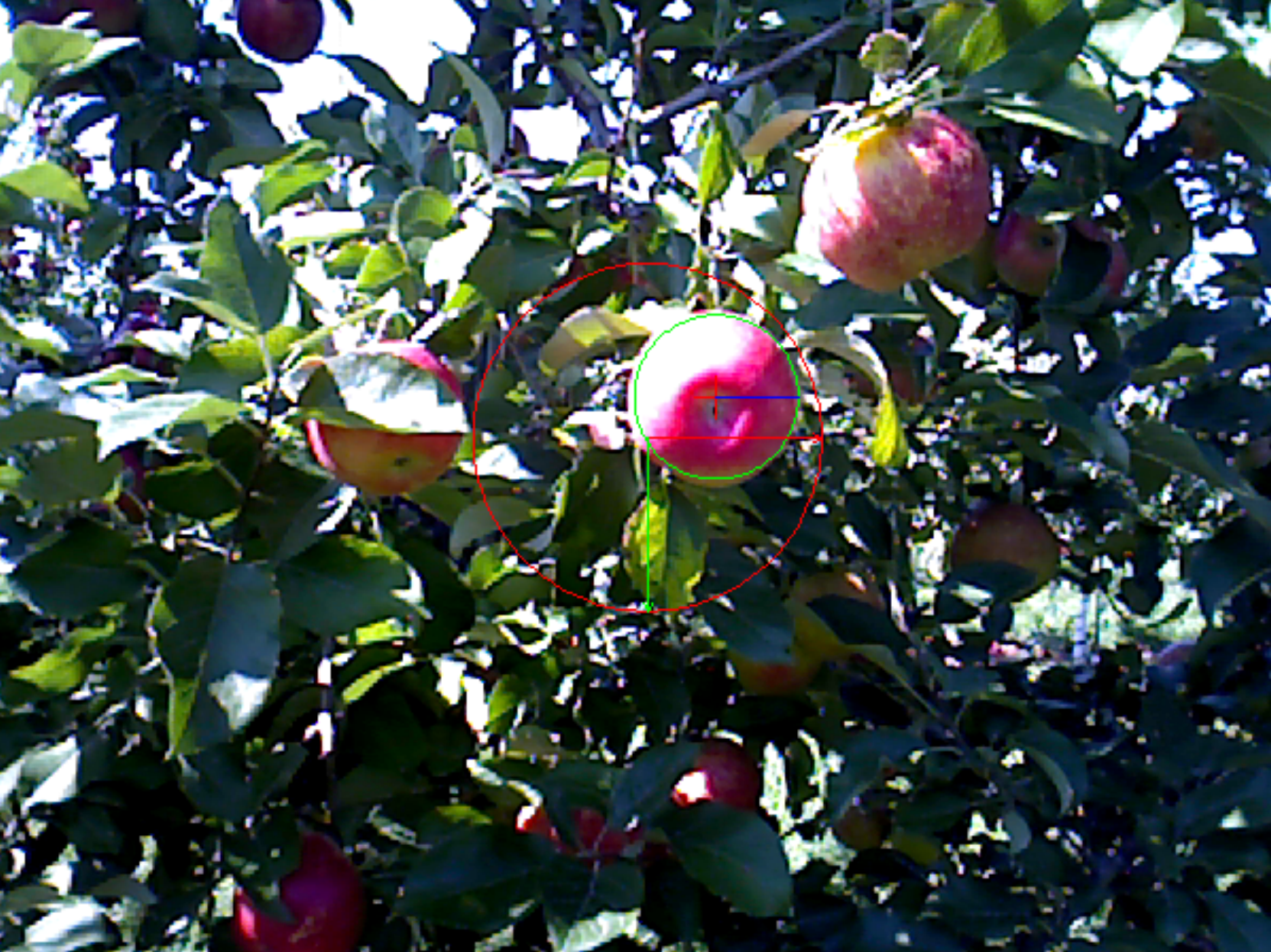}}
\subfigure[Converged position]{\label{fig:buttomUpB}\includegraphics[width=0.45\columnwidth]{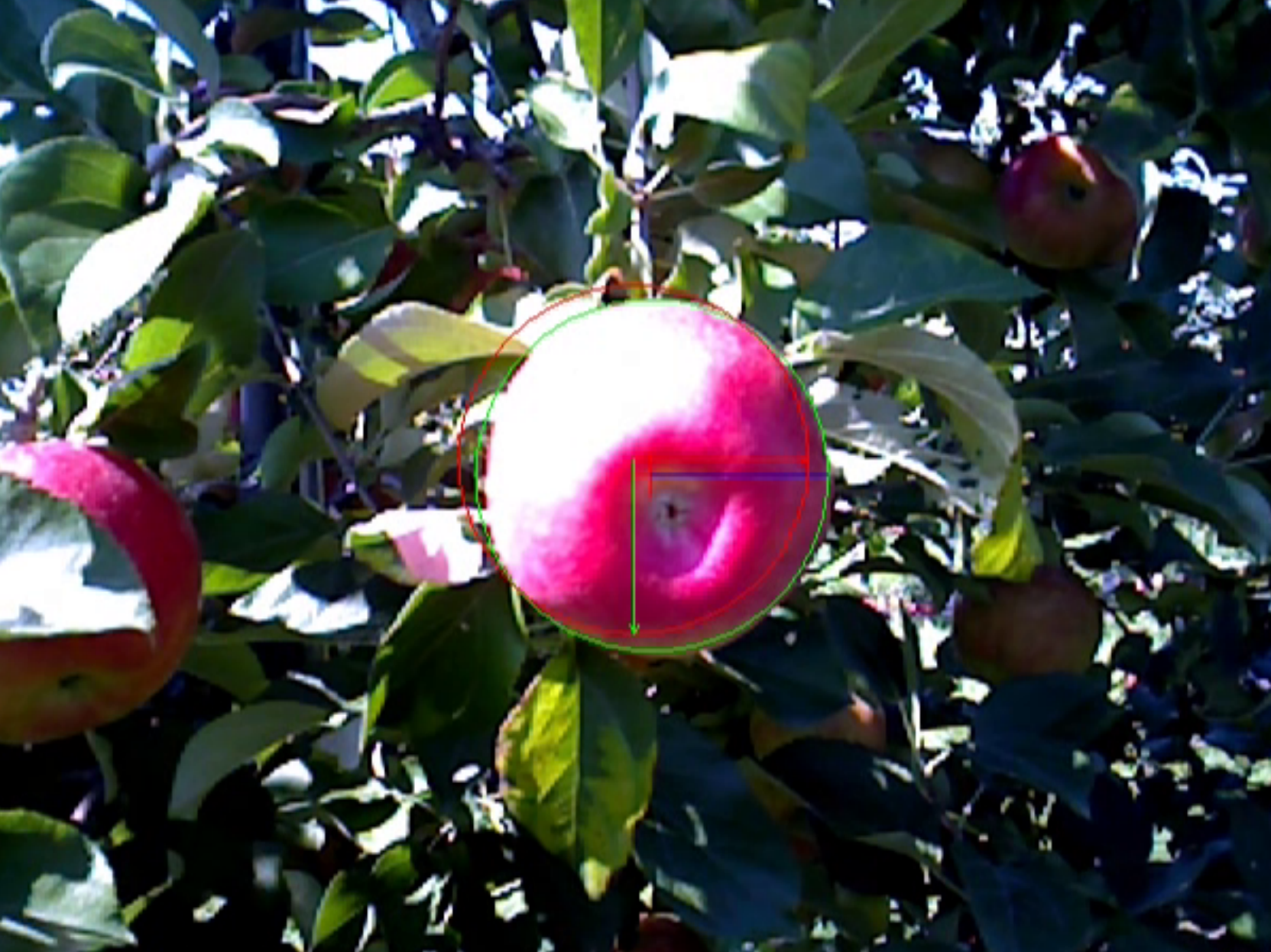}}

\subfigure[Velocity and error convergence]{\label{fig:buttomUpC}\includegraphics[width=\columnwidth]{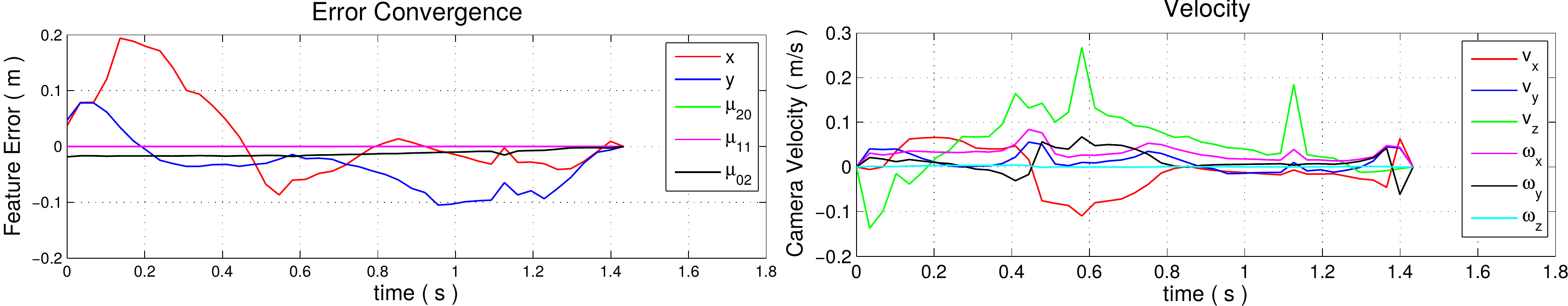}}
\caption{Servoing on apple in bottom up view \label{fig:buttomUp}}
\end{figure}

To test our approach from a different view point, we chose a ``bottom up configuration`` in which the camera looks up towards the blossoming point of an apple. Although approximately a third of the apple's surface is under the influence of strong specular reflections we show in Fig. \ref{fig:buttomUpA} that the apple can be identified and tracked correctly until convergence is achieved in Fig. \ref{fig:buttomUpB}. Fig. \ref{fig:buttomUpC} shows the convergence of the control law for this case. The oscillation of the controller is due to wind.

\subsection{Partial Occlusion Handling}
One of the problems in visual servoing is the handling of occlusions. We  differentiate between two different cases: (i)~Starting the visual servoing process with an occluded object in view and (ii)~The occurrence of occlusions during the process. The occurrence and avoidance of occlusions during the process of visual servoing have been well studied in recent years. For an overview see Cazy et al.~\cite{cazy_visual_????}. However in an outdoor setup like an apple orchard it is likely that already the initial view of a target object is occluded. 

\begin{figure}[htb]
\centering     
\subfigure[Initial position]{\label{fig:g}\includegraphics[width=0.45\columnwidth]{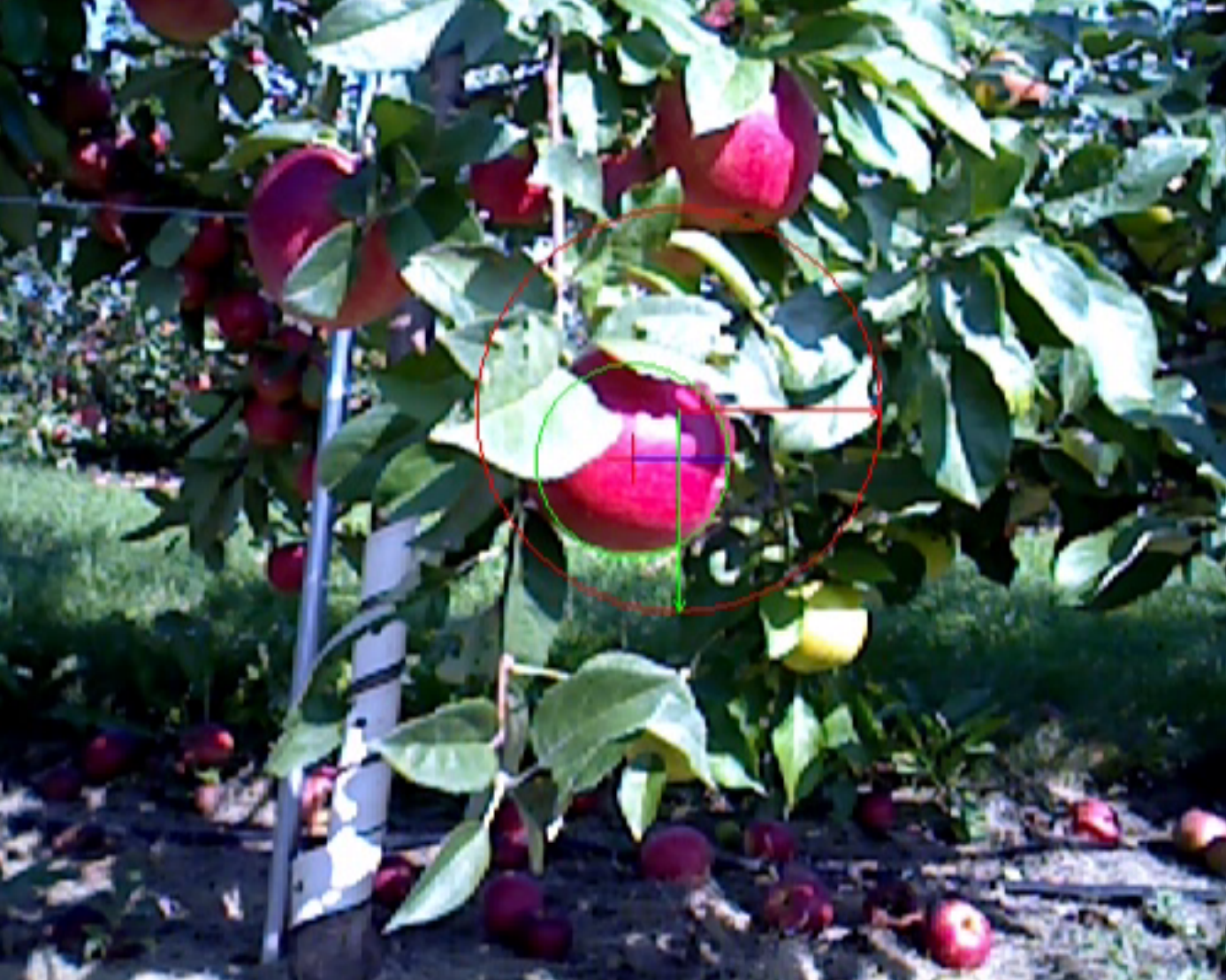}}
\subfigure[Converged position]{\label{fig:h}\includegraphics[width=0.45\columnwidth]{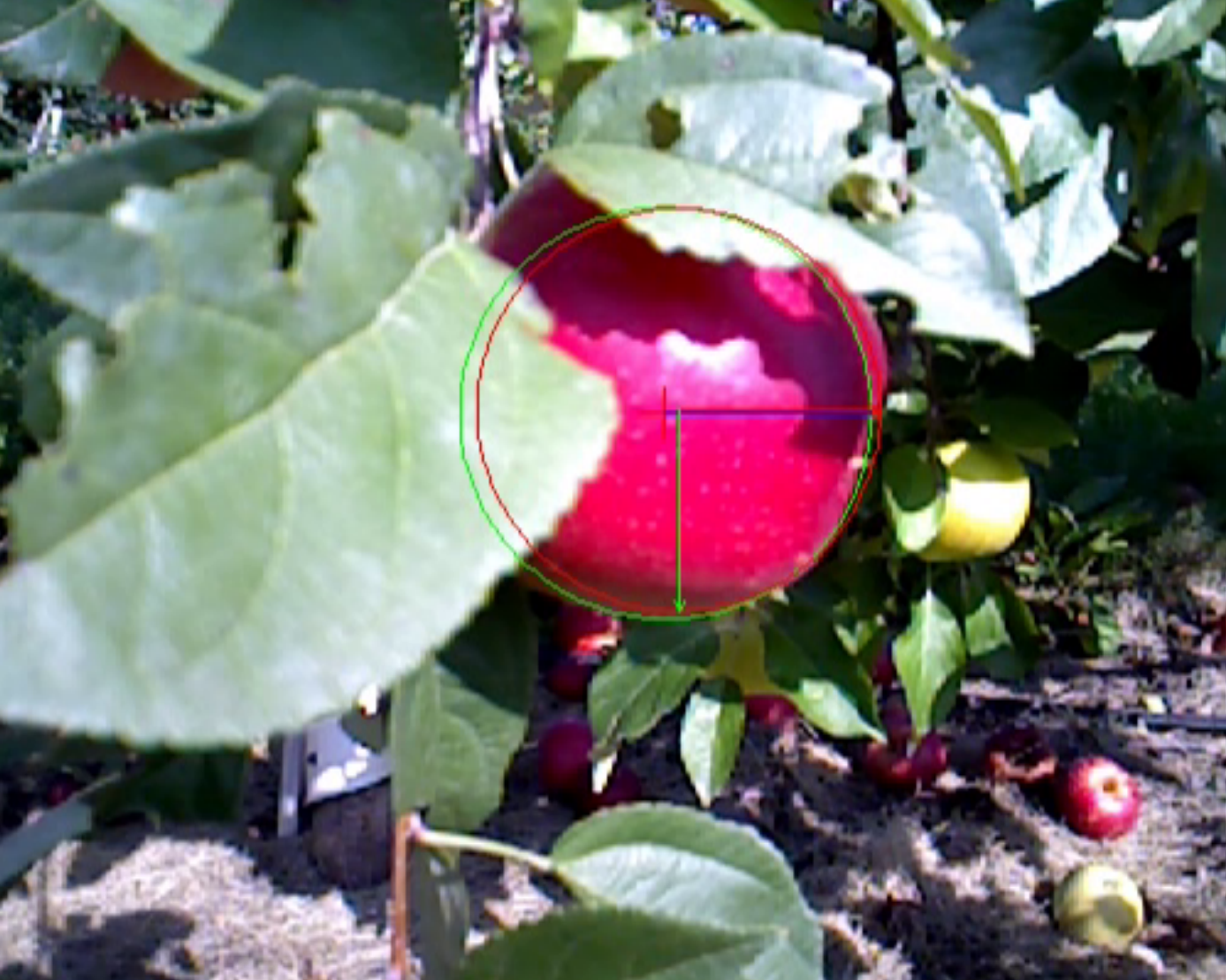}}

\subfigure[Velocity and error convergence]{\label{fig:i}\includegraphics[width=\columnwidth]{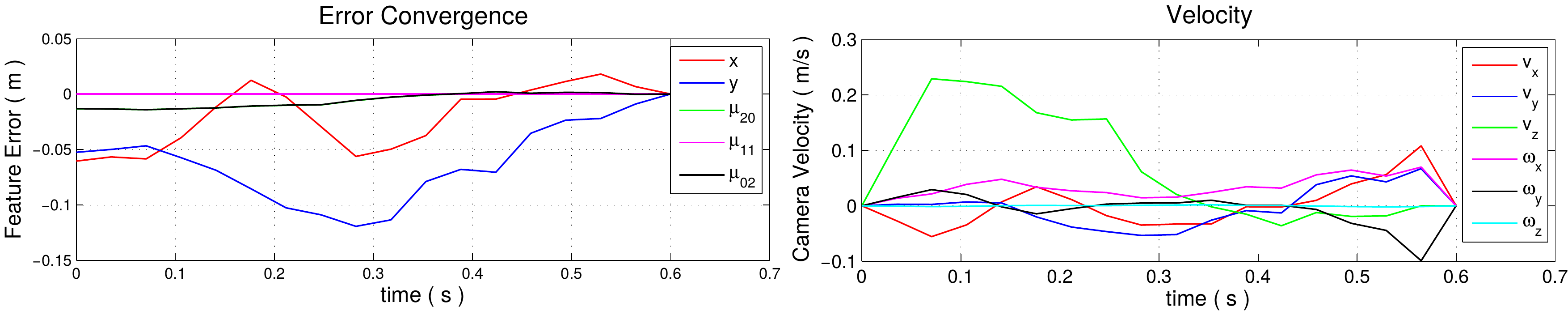}}
\caption{Outdoor experiment with inconsistent illumination, partial occlusion \label{fig:occlusion}}
\end{figure}

In this experiment the tracker has to handle partial occlusion as well as specular reflections and shadows. Although other apples are present in the image our tracker still focuses on the correct one thanks to the usage of decision matrix $D$. In Fig. \ref{fig:i} it can be seen  that convergence is achieved when the initial view contains a partially occluded target object. The tracker can not handle an arbitrary amount of occlusion. As soon as more than $50\mathrm{\%}$ of the apple boundary is occluded the tracker is either not able to detect the apple or keep it in view over the whole visual servoing distance. 

%% file: conc.tex
\section{CONCLUSION}
\label{sec:conclusion}

In this paper, we presented a robotic system for close-up inspection of fruits in orchards.  Specifically, we identified visual features and proposed a tracking mechanism, coupled with a servoing controller that we then tested in an orchard environment.
Our technique overcomes challenges associated with illumination changes, occlusions and disturbances. Controlled indoor experiments as well as outdoor experiments in an orchard on a windy day validated the effectiveness of the approach.
A direction for future research is to investigate the coupling of the controller with a mechanism to actively avoid occlusions. For this purpose it would be interesting to investigate the advantages of a stereo camera setup.
In cases where the target feature (e.g. the stem end) is not visible, manipulating the scene (e.g. by moving a leaf) is a challenging, yet exciting, research direction.